\pdfoutput=1

\documentclass[11pt]{article}
\PassOptionsToPackage{table}{xcolor}

\usepackage[final]{acl}

\usepackage{times}
\usepackage{latexsym}

\usepackage[T1]{fontenc}

\usepackage[utf8]{inputenc}

\usepackage{microtype}

\usepackage{inconsolata}
\usepackage{amsmath}
\usepackage{amssymb}

\usepackage{graphicx}
\usepackage{dsfont}
\usepackage{ragged2e}
\usepackage{float}
\usepackage{array}
\usepackage{tikz} 
\usepackage{wrapfig}
\usepackage{listings}
\usepackage[most]{tcolorbox}
\usepackage{xspace}
\usepackage{hyperref}
\usepackage[capitalize,noabbrev]{cleveref}
\usepackage{multirow}
\usepackage{booktabs}
\usepackage{svg}
\usepackage{tabularray}
\UseTblrLibrary{siunitx}
\usepackage{subcaption}
\usepackage{tabularx}
  \newcolumntype{C}{>{\centering\arraybackslash}X}
\definecolor{goodgreen}{RGB}{212,237,218}  %
\definecolor{badred}{RGB}{248,215,218} 
\definecolor{SBICtealDark}{HTML}{D64545}
\definecolor{ExactBlue}{HTML}{1874CD}

\newtcolorbox{quotebox}{
  enhanced,
  colback=SBICtealDark!15!white,
  colframe=SBICtealDark!40!black,
  coltitle=black,
  fonttitle=\sffamily\bfseries\large,  %
  fontupper=\itshape\rmfamily,   %
  left=4mm,
  right=4mm,
  top=2mm,
  bottom=2mm,
  boxrule=0.3pt,
  arc=2mm,
  drop shadow southeast,
  borderline west={2pt}{0pt}{SBICtealDark!70!black},
  before skip=12pt,
  after skip=12pt,
  boxsep=4pt
}

\newtcolorbox{promptbox}[1][]{
  enhanced,
  colback=ExactBlue!15!white,
  colframe=ExactBlue!40!black,
  coltitle=black,
  fonttitle=\bfseries\large\sffamily,
  fontupper=\ttfamily\small,
  left=5mm,
  right=5mm,
  top=3mm,
  bottom=3mm,
  boxrule=0.15pt,
  arc=1mm,
  borderline west={1pt}{0pt}{ExactBlue!80!black},
  before skip=12pt,
  after skip=12pt,
  title=\textcolor{white}{#1},
  enhanced jigsaw,
  overlay={
    \draw[ExactBlue!60!black, line width=0.3pt]
      (frame.north west) -- (frame.north east);
  },
}

\crefformat{section}{\S#2\color{blue}#1#3} %
\crefformat{subsection}{\S#2\color{blue}#1#3}
\crefformat{subsubsection}{\S#2#1#3}

\newcommand{\am}{\texttt{AutoMod}\xspace}

\definecolor{Maroon}{rgb}{0.501,0,0}

\title{\emph{Silencing Empowerment, Allowing Bigotry}: \\ 
Auditing the Moderation of Hate Speech on Twitch}

\author{
 \textbf{Prarabdh Shukla}\thanks{~~Equal contribution.}\textsuperscript{1},
 \textbf{Wei Yin Chong}\footnotemark[1]\textsuperscript{2},
 \textbf{Yash Patel}\footnotemark[1]\textsuperscript{1},
 \textbf{Brennan Schaffner\textsuperscript{2}},
\\
 \textbf{Danish Pruthi\textsuperscript{1}},
 \textbf{Arjun Bhagoji\textsuperscript{2, 3}}
\\
 \textsuperscript{1}Indian Institute of Science,
 \textsuperscript{2}University of Chicago,
 \textsuperscript{3}Indian Institute of Technology, Bombay
\\
 \small{
   \textbf{Correspondence:} \href{mailto:arjunp@iitb.ac.in }{arjunp@iitb.ac.in}
 }
}

\begin{document}
\maketitle

\begin{abstract}
\textit{\textcolor{Maroon}{\textbf{Warning}: This paper contains content that may be offensive or upsetting.}}
 
    To meet the 
    demands of content moderation, 
    online platforms 
    have resorted 
    to automated systems. 
    Newer forms of real-time engagement 
    (\emph{e.g.}, users commenting on live streams) on platforms like Twitch
    exert additional pressures 
    on the
    latency expected of such moderation 
    systems. 
    Despite 
    their prevalence,    
    relatively little is known about the effectiveness of these systems.
    In this paper, 
    we conduct an
    audit of Twitch's 
    automated moderation tool (\am) to 
    investigate its effectiveness in flagging hateful content. 
    For our audit,
    we create 
    streaming accounts 
    to act as siloed test beds, 
    and interface with the live chat using Twitch's APIs to send over $107,000$  comments collated from $4$ datasets.
    We measure \am's accuracy in flagging blatantly hateful content containing misogyny, racism, ableism and homophobia. Our experiments reveal that 
    a large fraction of hateful messages, up to $94\%$ on some datasets,
    \emph{bypass moderation}. 
    Contextual addition of slurs 
    to these messages 
    results in $100\%$ removal, 
    revealing \am's reliance on slurs as a moderation signal. 
    We also find that contrary to Twitch's community guidelines, \am blocks up to $89.5\%$ of benign examples that use sensitive words in pedagogical or empowering contexts. 
    Overall, our audit points to large gaps in \am's capabilities and underscores the importance for such systems to understand context effectively.\footnote{Our code and data can be found at \url{https://github.com/weiyinc11/HateSpeechModerationTwitch}}

    \end{abstract}

\section{Introduction}\label{sec:introduction}
\begin{figure}[t]
    \centering
    \includegraphics[width=0.8\columnwidth]{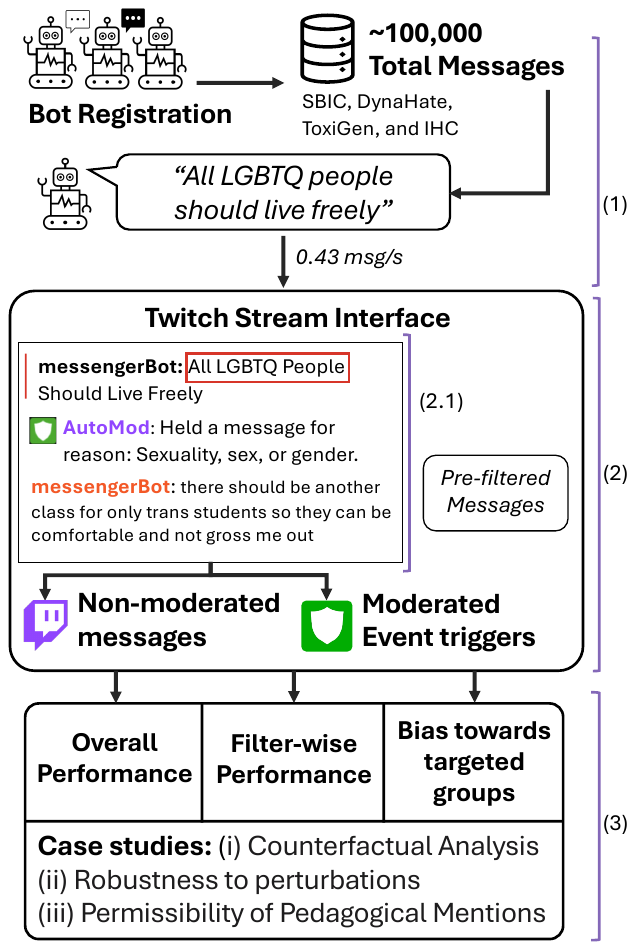}
     \caption{\textbf{The audit pipeline}: (\textbf{1}) Setting up bots to interact with \am and data collation, (\textbf{2}) Recording moderation decisions at scale, (\textbf{3}) Analysis of \am moderation decisions. (\textbf{2.1}) shows actual instances of moderated and non-moderated messages from our experiments in the Twitch interface. The moderation decisions clearly show the limitations of \am. 
    }
    \label{fig: experimental_design}
\end{figure}

For any online platform to exist without being overrun by hateful, pornographic, abusive, misogynistic and violent content, it must moderate what its users post \cite{gillespie2018custodians}.
Barring certain regions \cite{wiki:Network_Enforcement_Act}, 
there are few or no 
legal regulations 
dictating 
what is 
acceptable content 
on platforms
\cite{Schaffner_2024}. 
Moreover, in the Global North, platforms often enjoy legal protections from liability for hosting user-generated content \cite{uscode230, eu1214}.
This favorable regulatory framework bestows upon platforms the discretion to moderate content as they wish.
The discourse stands divided on the benefits of such freedom, with some crediting it for aiding the growth of online platforms~\cite{kosseff2019twenty}, while others critique it for enabling the proliferation of harmful content~\cite{wakabayashi2019legal}. Platforms codify 
the behavior expected of their users through  
terms of service and  community guidelines. 
While 
policies and guidelines may considerably vary across platforms,
most 
promise 
their users a safe space, 
free from online harm.
For instance, the community guidelines of Twitch, a video streaming platform, state:

\begin{quotebox}
    ``Twitch does not permit behavior that is motivated by hatred, prejudice or intolerance, including behavior that promotes or encourages discrimination, denigration, harassment, or violence based on the following protected characteristics: race, ethnicity, color,...
    We also provide certain protections for age.''
\end{quotebox}

The rhetoric notwithstanding, 
the pressures 
exerted
on moderation systems 
have never been higher. 
Platforms 
have increasingly begun to
integrate automated, usually machine learning-based, systems in their moderation pipelines \cite{gorwa2020algorithmic} due to the scale and velocity of content posted as well as the demanding latency expectations.\footnote{
From April 2022 to April 2023, social media platforms saw nearly $150$ million new users ($4.7$/second) \cite{searchenginejournalSocialMedia}.}
The practice of content moderation is further challenged by balancing the competing objectives of blocking broadly undesirable content and upholding users' freedom of expression. 
In response to these pressures, several platforms release aggregate data about the state of harmful content and corresponding actions taken~\cite{twitchtransparency24,twitterTransperencyReport,metatransparency}.  
Despite such transparency efforts, little is known about the underlying algorithms used for moderation and the biases they may introduce.

In this work, 
we conduct an
audit 
of Twitch's content moderation system. 
Twitch is a digital platform designed primarily for live streaming content which is created in channels or ``streams" that visitors can watch and interact with through a text-based live chat.
We focus on \emph{hate speech} content as it is  socially important and easier to define than other harmful content such as misinformation~\cite{Schaffner_2024}.
We identify 
Twitch as a fertile
platform for auditing 
due to three key advantages it offers: (i)
Twitch is widely used \cite{twitchtracker}; 
(ii) streams on Twitch can be ``siloed,'' allowing us to set up controlled experiments
where only the research team
can view the content to be tested (iii) the platform 
provides streamers 
with a suite of machine learning-based moderation tools collectively called \am \cite{twitchautomod}, which  
provides streamers control over different categories of harmful content. Within each category, 
it offers 
options 
to block harmful content targeted at different identity groups,
with knobs to
vary the
extent of moderation (\cref{sec:twitch}).
Through this audit, we seek to answer the following 
\emph{key research questions} (\cref{sec: eval}): 
(i) How effective is \am at flagging commentary rife with hate?; (ii) How specific and effective are individual filters at blocking hate of different kinds and intents?; and (iii) Are  moderation rates 
consistent across 
different target groups, 
or are certain
groups 
disproportionately 
affected? We use case-studies (\cref{sec:ablations}) to nuance these by asking (i) how reliant are the moderation filters on explicit slurs; (ii)
how do they
respond to 
sensitive but 
pedagogical content 
about marginalized groups?; and (ii)
how 
far can a filter-aware
malicious actor 
bring down 
their performance? 
To answer such questions, we develop a framework that allows us to stress test \am at scale by launching chatbots in a siloed sandbox.
For our study, we use four datasets -- a real world comment dataset (SBIC), a real world implicit hate dataset (IHC), a synthetic dataset on implicit hate (ToxiGen) and a synthetic dataset designed to fool hate classifiers (DynaHate). We discuss the hate speech datasets used in \cref{subsec: data} and our experimental setup in \cref{subsec: setup}.

Our audit 
reveals that Twitch's
automated moderation
is far from adequate (\cref{sec:results}), 
flagging only $22\%$ of
hateful content
even at its most stringent setting. 
We observe that hateful samples concerning \textit{Race, Ethnicity and Religion}
are least flagged, with only $12.3\%$ of hateful examples being caught. Further, on some datasets, we find that an overwhelming $98\%$ of hate 
targeted at \textit{Mentally Disabled folks} escapes moderation. 
On the two implicit hate datasets we use, \am is only able to flag $6.8\%$ of hateful examples, implying that the filters rely heavily on slur-based moderation and miss out on implicit hate.
Worryingly,
empowering or positive phrases about communities (\emph{e.g.} \cref{fig: experimental_design}) are flagged, and we find \am to be quite brittle to semantic-preserving perturbations (See \cref{sec:ablations}).

While it is undeniable that Twitch provides powerful, customizable tools for moderation to its users, our audit serves as an important reminder that these tools must be comprehensively tested and their limitations made clear. Third-party audits like ours can
inform the 
discourse on content moderation
by grounding the 
discussion with quantitative evidence regarding the challenge of moderating in a holistic fashion. We hope our methodology inspires and is generalized to audit automated content moderation and other decision-making systems across platforms and data modalities.

\section{Related Work}

\paragraph{Content Moderation} \label{sec:cont_mod_bg}
Given its importance, content moderation has been studied extensively~\cite{keller2020facts}. 
Prior work has systematically analyzed and critiqued the moderation policies from various online platforms, either focusing on a single platform~\cite{chandrasekharan_internets_2018, fiesler+al:2018, Brian_keegan_2017} or on a industry-wide analysis across many platforms~\cite{Schaffner_2024}.
Other work has focused on specific aspects of moderation, such as user reactions to moderation~\cite{Cai_2024, ribeiro2023automatedcontentmoderationincreases}, the effects moderation has on community behavior~\cite{Chancellor:2016:TIC:2818048.2819963, chandrasekharan2017you, Chang-Recidivism:19}, and the disproportionate negative effects of moderation on blind users \cite{Lyu_2024}.
Prior work has also proposed using various AI models to assist in automated content moderation  \cite{kumar2024watchlanguageinvestigatingcontent, francoLLMContentMod, kollaLLMMod, grayMLCopyRight}. For an in-depth discussion of current content moderation efforts, we refer the reader to \cite{arora2023detectingharmfulcontentonline}. 
Our work complements existing work by auditing real-world automated content moderation algorithms.

\paragraph{Auditing Algorithms} 
Auditing, as defined by \cite{gaddis2018audit}, is a methodology used to deploy randomized controlled experiments in a field setting. When audits target algorithms and computer systems---termed ``algorithm audits''~\cite{Sandvig2014AuditingA, metaxaAlgoAudit}---the outputs of a system are analyzed when making minor changes to the input, which can lead to insights about the system as a whole.
Most often, algorithm audits investigate underlying biases and discriminatory behavior of a system~\cite{edelmanAirbnb, pmlr-v81-speicher18a, chenGenderSearchEng, metaxa2021image}. 
Other studies have audited whether platforms enforce their policies effectively and fairly, targeting, for example, the political advertising policies on Facebook and Google~\cite{pochatFacebookAdAudit,matias2021softwaresupportedauditsdecisionmakingsystems}.
Algorithm audits  
span a wide variety of domains examining several platforms, 
including housing (rental platforms such as AirBnb) \cite{edelmanAirbnb}, 
ride sharing (Uber)~\cite{chen2015peeking},
healthcare \cite{obermeyer2019dissecting}, 
employment and hiring \cite{chen2018investigating, pmlr-v81-speicher18a}, 
advertisements \cite{pmlr-v81-speicher18a},
and
product pricing~\cite{mikians2012detecting}. In a typical 
algorithm audit such as ours,
auditors work with only
black-box 
access to the system,  
and need to draw conclusions 
with just that level of access~\cite{cen2024transparencyaccountabilitybackdiscussion}. A concurrently published study \cite{Hartmann_2025} evaluates various moderation APIs (such as OpenAI's moderation API \cite{markov2023holisticapproachundesiredcontent}) that have recently emerged. The choice of datasets and methodology in this study is similar to ours, further validating our approach. \citep{Hartmann_2025} note that most of the evaluated APIs under-moderate implicit hate and over-moderate pedagogical/empowering content, reclaimed slurs, and counter-speech. They also highlight that moderation APIs often rely on group identity keywords such as “black” when making moderation decisions. We make almost identical observations for \am (See \cref{sec:results}, \cref{sec:ablations}) and together, these results point to the overall gap that exists between the capabilities of SoTA language technology and that used for commercial moderation (See Figure \ref{fig:sota_comparison}).

\paragraph{Live Streaming on Twitch}

With Twitch's rise in popularity, it has been the subject of several recent studies, including as an emergent political space by modeling the roles of different actor groups \cite{bravoTwitchPoliticization}. Another work studies volunteer moderators on Twitch by analyzing their recruitment, motivation and roles in comparison to other online platforms \cite{seeringTwitchVolunteerMod}. Recent work has revealed that waves of attacks (termed ``hate raids" by popular media) which were experienced across Twitch in 2021 were targeted on the basis of creator demographics \cite{han2023hateraidstwitchechoes}. To the best of our knowledge, we are the first to study automated content moderation on Twitch.

\section{Methodology}\label{sec:Methodology}

In this section, we first describe our survey of online platforms. 
We then describe the datasets we use for the audit, followed by a mathematical description of the audit.

  \subsection{Exploring Platforms}\label{sec:twitch}
 We set two requirements for choosing a platform to audit: (i) \textit{Harm reduction}: the content posted as a part of the audit should only be visible to a controlled group (\textit{i.e.}, ``siloed"); and (ii) \textit{Advanced moderation tools}: a suite of configurable moderation tools (preferably leveraging native machine learning models).  
 The first requirement stems from the ethical requirement to minimize harm to unsuspecting users and not contribute to the existing deluge of hate speech~\cite{twitchtransparency24}.
ML systems tend to have biases and often have characterizable faults, which makes it important to audit a platform that uses this technology, hence the second requirement.
 Moreover, while some platforms may employ less advanced moderation systems in the form of blocklists, 
 auditing such systems would only test the comprehensiveness of the lists, not how well the provided tools work.
We surveyed the 43 largest platforms that host user-generated content \cite{Schaffner_2024} to check if any meet our requirements, finding two suitable platforms---Reddit and Twitch. We also separately considered Discord as a candidate platform, but we ultimately chose Twitch for this audit due to its moderation system being particularly configurable and also revealing moderation justifications. 
We compare these 
three platforms in light of our requirements (Table \ref{tab: platform_survey} in~\cref{appsec: twitch_details}).

\paragraph{What moderation tools does Twitch offer?}
There are three main kinds of interactions on Twitch: streamer-to-viewer, viewer-to-streamer, and viewer-to-viewer, with content being moderated across these interactions.   
The moderation 
at the streamer-to-viewer level happens via machine learning-based tools 
that check the audiovisual stream for content that violates Twitch's platform-wide policies.
Auditing moderation of audiovisual content
is beyond the scope of this paper 
but presents an interesting direction for future work (\cref{sec:discussion}). 
In this audit, we focus on moderation of the viewer-to-streamer and viewer-to-viewer interactions, which happen via the live chat interface. 

Streamers, as well as designated moderators, can configure Twitch's \am tool to handle the potentially high volume of chat content automatically with filters for various types of unwanted content. \am uses machine learning~\cite{twitchautomod} to detect the unwanted content and surface it in the Moderator view (\cref{fig:moderatorView}). In addition, streamers can also populate a blocklist of specific words or phrases to restrict from the chat. 
\am's customisability to vary its detection in levels of strength and content categories makes it an interesting testbed for auditing. For instance, detection sensitivity for \textit{Profanity} and \textit{Discriminatory Content} can be independently set on $5$-point scales (See \cref{subsec: setup} for details on how we configured \am for our audit study.) 
Twitch also allows viewers themselves (independent of the streamer) to toggle Chat Filters (\cref{fig:chatFilter}) to block certain content categories, but these filters are not as granular as those offered to moderators. A Smart Detection option is also in beta, which allows moderation rules to be learned from moderation actions. Since this requires manual moderation, it is beyond the scope of this paper.

  \subsection{Scope of harmful content: Hate Speech}\label{subsec: data}
Amongst
the
categories of problematic
content that Twitch moderates, 
the majority of moderated content falls into the Sexual Conduct, Harassment, and Hateful Conduct categories (as per Twitch's 2024 Transparency Report~\cite{twitchtransparency24}).
We focus 
our audit on hateful speech, 
which more specifically 
falls under the 
\emph{Discrimination \& Slurs} category of Twitch's moderation (Table~\ref{tab: automod_categories}). 
Speech in 
 this category is often nuanced in its intent and use 
 of language, making for an interesting study. 
 We use four datasets for our experiments: DynaHate \cite{vidgen-etal-2021-learning}, Social Bias Inference Corpus (SBIC) \cite{sap-etal-2020-social}, ToxiGen \cite{hartvigsen-etal-2022-toxigen} and the Implicit Hate Corpus (IHC) \cite{elsherief-etal-2021-latent}, of which DynaHate and ToxiGen are synthetically generated. Of these, the first three datasets come with annotations specifying the target groups, allowing us to stratify our analysis (\cref{sec:results}).\footnote{We did not find enough examples in IHC for each category, and therefore did not use it for the stratified analysis.} For SBIC, each example comes with an offensiveness score ranging from 0 to 1, where 0 represents the least offensive rating and 1 represents the most offensive rating. We use a threshold of 1 on the offensiveness score to obtain ground truth labels unless specified otherwise. A more detailed, dataset-wise description is provided in \cref{app:data}.

\subsection{Audit Design}\label{subsec: notation}
\noindent \textbf{Notation and Terminology:} We define a black box content moderation system (such as \am) as a two-tuple $\mathcal{S} = ({\mathcal{F}}, \mathcal{C})$. Here, $\mathcal{F} = \{ \mathcal{F}_1, \dots, \mathcal{F}_k\}$ is a set of $k$ black-box moderation functions (filters) and $\mathcal{C} = \{c_1, \dots, c_k\}$ is the corresponding set of $k$ abstract criteria that map on to real-world concepts such as disability or misogyny, usually derived from the policy that $\mathcal{S}$ aims to enforce. Each $\mathcal{F}_i : T \to \{0,1\}$ is a function that maps a text $t\sim T$ to either label $0$ (benign) or $1$ (violation) based on criterion $c_i$. Here $T$ represents the distribution of all possible textual inputs to the moderation system.
The moderation decision for an input $t$ is determined by the \emph{active moderation function} $\mathcal{F}_A$ which corresponds to the 
set of active filters $\mathcal{C}_A$. $\mathcal{C}_A$ is the set of criteria according to which content needs to be moderated. 
In \am, each filter $\mathcal{F}_i$ is further parameterized by $\alpha$, the \textit{filtering level}---which is a discrete measure that modulates
the strictness of enforcement.

\paragraph{Filter Choices:} In automated content moderation, the elements of the set $\mathcal{C}$ frame the moderation policy of the platform as a whole, while $\mathcal{C}_A$ represents the decisions made by the current streamer or moderator. When we say a particular set $\mathcal{C}_A$ of filters is ``turned on,'' we are referring to the active moderation function $\mathcal{F}_A$ being $\mathds{1}(\bigcup\limits_{c_i\in \mathcal{C}_A} \mathcal{F}_{i} = 1 )$ which returns $1$ when any of the filters returns $1$. In this study, we focus on auditing moderation functions of a subset $\tilde{\mathcal{C}}= \{\text{Disability, SSG, Misogyny, RER}\}$\footnote{SSG stands for \textit{Sex, Sexuality and Gender} and RER stands for \textit{Race, Ethnicity and Religion}.}, corresponding to the broad category of \emph{Discrimination and Slurs} (\cref{subsec: data}). Our analysis in~\cref{sec:results} proceeds by varying $\mathcal{C}_A \subseteq \tilde{\mathcal{C}}$ (and correspondingly $\mathcal{F}_A$) with respect to different subsets of our base dataset $\mathcal{D}$. 

In our experiments we construct $\mathcal{D}= \left( \bigcup_{c_i \in \tilde{\mathcal{C}}} D_{c_i} \right) \bigcup  D_{\text{benign}}$, where each text sample in $D_{c_i}$ has \emph{at least one ground-truth label mapping to $c_i$}. $D_{\text{benign}}$ are text samples that do not correspond to any category in $\mathcal{C}$, the set of criteria Twitch allows filtering for.
Given the ground-truth labels, we use standard metrics such as precision and recall to measure the  effectiveness of the filters at detecting hate speech for the overall dataset, for different categories, and aimed at different target groups within each category.

\paragraph{Filter- and Policy-Informed Text Generation:} The base datasets we use contain generic examples of speech, hateful or not, which are not tailored to investigate the robustness of \am's moderation functions and the alignment between Twitch's stated policies and \am operation. In~\cref{sec:ablations}, we further test $\mathcal{F}$ with bespoke constructed samples for implicit hate detection, context-awareness and robustness to semantic-preserving inputs.

\section{Experiments \& Results}\label{sec: eval}

In \cref{subsec: setup} we describe the experimental setup that allows us to record moderation decisions from Twitch at scale. We then discuss \am's performance in different contexts in \cref{sec:results}. 
\subsection{Setup}\label{subsec: setup}
Here, we describe the setup and components of our experimental pipeline (\cref{fig: experimental_design}) in brief (details in \cref{appsec: setup}). For simplicity, unless otherwise indicated, the level for each filter was set to $\alpha=4$, which is `Maximum Filtering'. 
Twitch provides functionality for registering and developing chatbots, typically used by streamers to facilitate stream engagement, such as viewer rewards or stream donations.
To send messages programmatically and at scale, we created chatbots that we registered with Twitch's Developer Console using a combination of online burner phone numbers, personal phone numbers, and alias email accounts. 
Running one instance of the experimental pipeline requires three authenticated bots: a messenger bot, a receiver bot, and a Pubsub bot.

We use the messenger bot to send messages to the chat stream while adhering to Twitch's Chat Rate Limits~\cite{TwitchChatDoc} to prevent message duplication or omission. 
We then use a receiver bot to mimic a user viewing the chat stream---and record all messages that were \textit{not} moderated. To observe the moderated messages, we subscribe the third bot to Twitch's Pubsub events, which provide information about the problematic fragments that leads to moderation as well as internal categorical labels such as \texttt{Ableism}, \texttt{Misogyny}, \texttt{Racism}, and \texttt{Homophobia} (\cref{fig:pubsub_result}). While these categories are not detailed in the \am documentation, we infer an injective relationship between the content categories in the \am documentation and internal categories from the API usage. With these labels, we are able to investigate \am's stated reasons for moderation and conduct further category-specific analysis in~\cref{subsec: level2}. In total, we send and record the moderation (in)activity for around $300,000$ messages from December 2024 to January 2025 for our experiments.  
\paragraph{Challenges Faced:} 
Despite adhering to the rate limit, we speculate that Twitch's safety measures against fraudulent activity \cite{twitchtransparency24}
led to the repeated banning of our chatbots, which forced us to create $30$ different developer accounts. 
Additionally, we observe a \emph{third category of messages} that were not detected by either the receiver or the pubsub bot, indicating a third undocumented destination for messages on Twitch. We suspect that these messages---given that most of them contained specific hateful slurs---are caught at the Service Level in Twitch's moderation pyramid (See \cref{subfig:mode_pyramid}) and therefore filtered out \emph{prior to} \am which occurs at the Channel Level. We refer to such messages as \textit{pre-filtered} from here on and discuss their effects later in this section.

\subsection{Results}\label{sec:results}
\begin{table*}[t]
\centering
\begin{tabular}{@{}lcccccc@{}}
\toprule
\textbf{Dataset} & \multicolumn{1}{l}{\textbf{Accuracy}} & \multicolumn{1}{l}{\textbf{Precision}} & \multicolumn{1}{l}{\textbf{Recall}} & \multicolumn{1}{l}{\textbf{TNR}} & \multicolumn{1}{l}{$\mathbf{F1_{\text{P,R}}}$} & \multicolumn{1}{l}{$\mathbf{F1_{\text{TPR,TNR}}}$} \\ \midrule
SBIC \cite{sap-etal-2020-social}             & \cellcolor{goodgreen} 0.73                                  & \cellcolor{badred}0.42                                   & 0.19                                & 0.91                             & 0.26                               & 0.31                                                                \\
DynaHate* \cite{vidgen2021learning}        & \cellcolor{badred} 0.49                                  & 0.54                                   & \cellcolor{goodgreen} 0.41                                & \cellcolor{badred} 0.59                             & \cellcolor{goodgreen}0.47                               & \cellcolor{goodgreen}0.48                                                                \\ 
ToxiGen*  \cite{hartvigsen-etal-2022-toxigen}       & 0.53                                  & \cellcolor{goodgreen}0.86                                   & 0.07                                & \cellcolor{goodgreen} 0.98                             & 0.13                               & 0.13                                                                \\ 
 IHC \cite{elsherief-etal-2021-latent}       & 0.52                                  & 0.70                                 & \cellcolor{badred} 0.06                                & 0.97                             &\cellcolor{badred} 0.12                               & \cellcolor{badred}0.11                                                                \\ \hline
Overall          &    0.55                               &  0.56                                  &    0.22                             &  0.84                            &  0.32                              &  0.35                                                               \\ \bottomrule
\end{tabular}
\caption{\textbf{\am's aggregate performance}. 
\am is reasonably accurate with benign examples, but struggles with the hateful ones. 
Recall on ToxiGen and IHC 
datasets is lower
than the other datasets implying \am is weak at detecting implicit hate. 
(*) is used to indicate synthetic data. 
\textcolor[rgb]{0.16,0.66,0.27}{Green} indicates the best value and \textcolor[rgb]{0.86,0.21,0.27}{Red} indicates the worst value of a metric.}
\label{tab:exp_l1}
\end{table*}

In this section, we first present analysis on the overall effectiveness of \am. We then discuss filter-wise performance and filter-specificity, followed by target group specific results.

\begin{figure}[h]
    \centering
    \includegraphics[width=1\columnwidth]{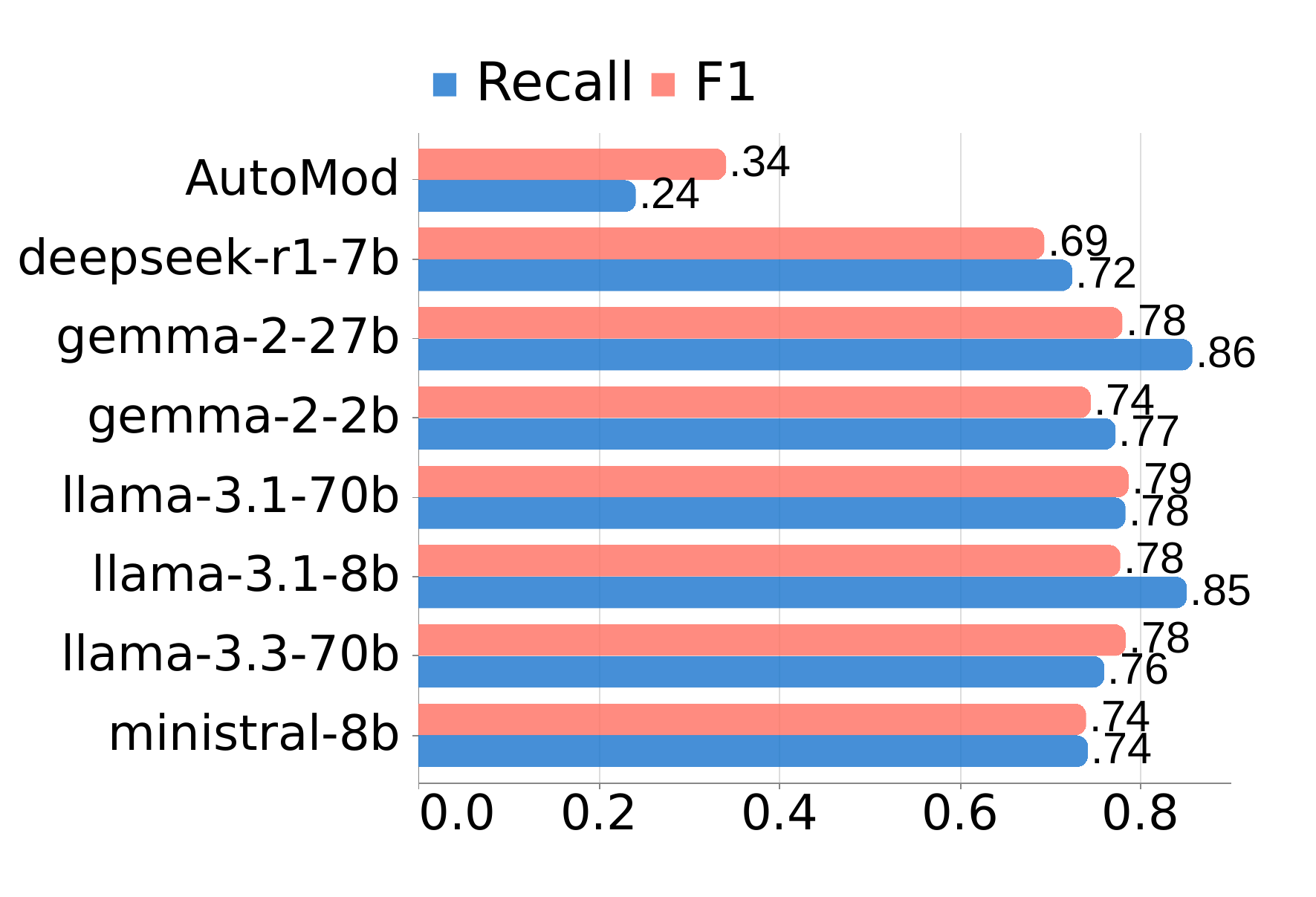}
    \caption{\textbf{Comparison of \am to State-of-The-Art (SoTA) language models}. Language models were prompted with a zero-shot instruction containing Twitch's community guidelines. For experimental setup, see \cref{app:benchmarking}.}
    
    \label{fig:sota_comparison}
\end{figure}

\paragraph{\am's Overall performance:}\label{subsec: level1}
We pass hateful content to Twitch as described in~\cref{subsec: setup} and obtain, corresponding to each example, a binary label $Y \in \{0,1\}$ that indicates if the example was moderated ($Y=1$) or not ($Y=0$). Here, $\mathcal{C}_A=\tilde{\mathcal{C}}$, implying all categories under \emph{Discrimination and Slurs} are considered. We then compare these to the ground-truth labels and measure the accuracy, precision, recall or True Positive Rate (TPR), and F1 score. 
As described in \cref{sec:introduction}, a content moderation system aims to strike a balance between the competing objectives of i) blocking hateful content while ii) safeguarding freedom of speech. $F1_{\text{TPR, TNR}}$ 
is the harmonic mean between the TPR and the True Negative Rate (TNR). It measures the simultaneous optimization of both objectives. 
Recall captures performance on objective i). \am achieves only $22\%$ recall overall which is quite low when compared to open source classifiers trained on similar data~\cite{sap-etal-2020-social} and also when compared to zero-shot performance of SoTA language models (see Figure \ref{fig:sota_comparison}). \am struggles on even the most evidently hateful real-world data, flagging just $19$\% of content that SBIC annotators unanimously labeled as hateful. On the two implicit hate datasets, ToxiGen and IHC, we observe a much lower recall ($6$ and $7\%$ respectively) than the others which indicates that \am is poor at recognizing implicit hate and lacks understanding of context (we further validate this in \cref{sec:ablations}).  We also evaluate Twitch's Chat Filter (described in \cref{sec:twitch}) and our findings (see \cref{app:chat_filter}) reveal that it behaves identically to \am at $\alpha = 4$. Thus, we focus on \am for the rest of the paper.

\textit{Offensiveness thresholds:} For SBIC, we used different thresholds on the offensiveness score to obtain ground truth labels. A higher threshold reflects stricter human agreement for categorizing an example as hateful. As we relax the threshold on the offensiveness score, accounting for varied opinions and more nuanced instances of hate, the recall drops further (\cref{fig:recall_threshold}). Further results at different thresholds are in~\cref{appsec: sbic_threshold}. 

\textit{FN/FP analysis:} We speculate that the high FNR ($1-~$Recall) is primarily due to implicit hate. To validate this, we use an LLM to identify swear words in the false negatives (see \cref{appsec:fnfp}) and find that $89.8\%$ of False Negatives are devoid of profanity (\emph{i.e.}, are implicitly hateful). We also note that \am performs well with the negative class, as is evidenced by the high TNR. 
Using a similar analysis on DynaHate --- which has a relatively higher FPR --- we find that nearly $73\%$ of false positives contain profanity. This analysis hints at \am's heavy reliance on using swear words as a signal for hate (further details in \cref{sec:ablations}). 

\textit{Conversation level context awareness:} While the analysis of false positives and false negatives hint at very limited context-awareness at the message level, we conducted another experiment to ascertain if \am possesses any context-awareness at the conversation level (\textit{i.e.}, across different messages passed to the chat). For this, we selected $100$ hateful and $100$ benign examples from SBIC. We first passed the 100 hate examples alone and observed that $35$ were blocked. Next, we interleaved the same $100$ hateful examples with the $100$ benign examples and observed that the same $35$ messages were blocked again, indicating that \am does not possess conversation-level context awareness. In addition, we also tested our framework with different permutations of examples to ascertain the reproducibility of our results. We observed that changing the order of input messages does not change the moderation decision for any example.

\begin{table*}[t]
\centering
\begin{tabular}{@{}lrrrrrrrr@{}}
\toprule
\multicolumn{1}{c}{\textbf{Dataset}} & \multicolumn{2}{c}{\textbf{SBIC}} & \multicolumn{2}{c}{\textbf{DynaHate}} & \multicolumn{2}{c}{\textbf{ToxiGen}} & \multicolumn{2}{c}{\textbf{Overall}} \\ \midrule
\textbf{Filter} & \multicolumn{1}{r}{\textbf{$\mathcal{R}$ (\%)}} & \multicolumn{1}{r}{\textbf{\textit{Pf} (\%)}} & \multicolumn{1}{r}{\textbf{$\mathcal{R}$ (\%)}} & \multicolumn{1}{r}{\textbf{\textit{Pf} (\%)}} & \multicolumn{1}{r}{\textbf{$\mathcal{R}$ (\%)}} & \multicolumn{1}{r}{\textbf{\textit{Pf} (\%)}} & \multicolumn{1}{r}{\textbf{$\mathcal{R}$ (\%)}} & \multicolumn{1}{r}{\textbf{\textit{Pf} (\%)}} \\ \midrule
Disability & 22.4 & 2.0 & 44.2 & 4.4 & 2.9 & 13.8 & 10.6 & 6.1 \\
Misogyny & 27.3 & 0.8 & 25.8 & 1.4 & 3.9 & 4.6 & 19.0 & 1.5 \\
RER & 17.3 & 36.1 & 20.5 & 18.8 & 6.6 & 21.5 & 12.3 & 22.0 \\
SSG & 32.0 & 64.1 & 25.3 & 55.3 & 5.7 & 44.3 & 17.5 & 54.8 \\ \bottomrule
\end{tabular}
\caption{\textbf{Filter-wise recall across datasets}. We report recall ($\mathcal{R}$) and the percentage of examples that were pre-filtered (\textit{Pf}). The \textit{Misogyny} filter is the most effective. The SSG filter relies on pre-filtering more heavily than other filters.}
\label{tab:exp_l2_relevant}
\end{table*}

\paragraph{Filter specificity and effectiveness:} \label{subsec: level2}
We manually categorize hateful examples according to the four filters we analyze:  Disability, Misogyny, RER and SSG, thus constructing $4$ subsets $\mathcal{D}_{c_i} \subset \mathcal{D}$ (See Appendix \ref{app:label_mapping}). Quality control analysis on our data subsets is provided in Appendix \ref{app:data_quality_control}. The primary objective of this analysis is to assess how well a filter $\mathcal{F}_i$ models the criterion $c_i$ associated with it. Note that we receive \am's internal categories for any moderated message, allowing us to determine which filter led to moderation even when multiple are on.

\begin{figure*}[t]
       \includegraphics[width=\textwidth, height=0.5\textwidth]{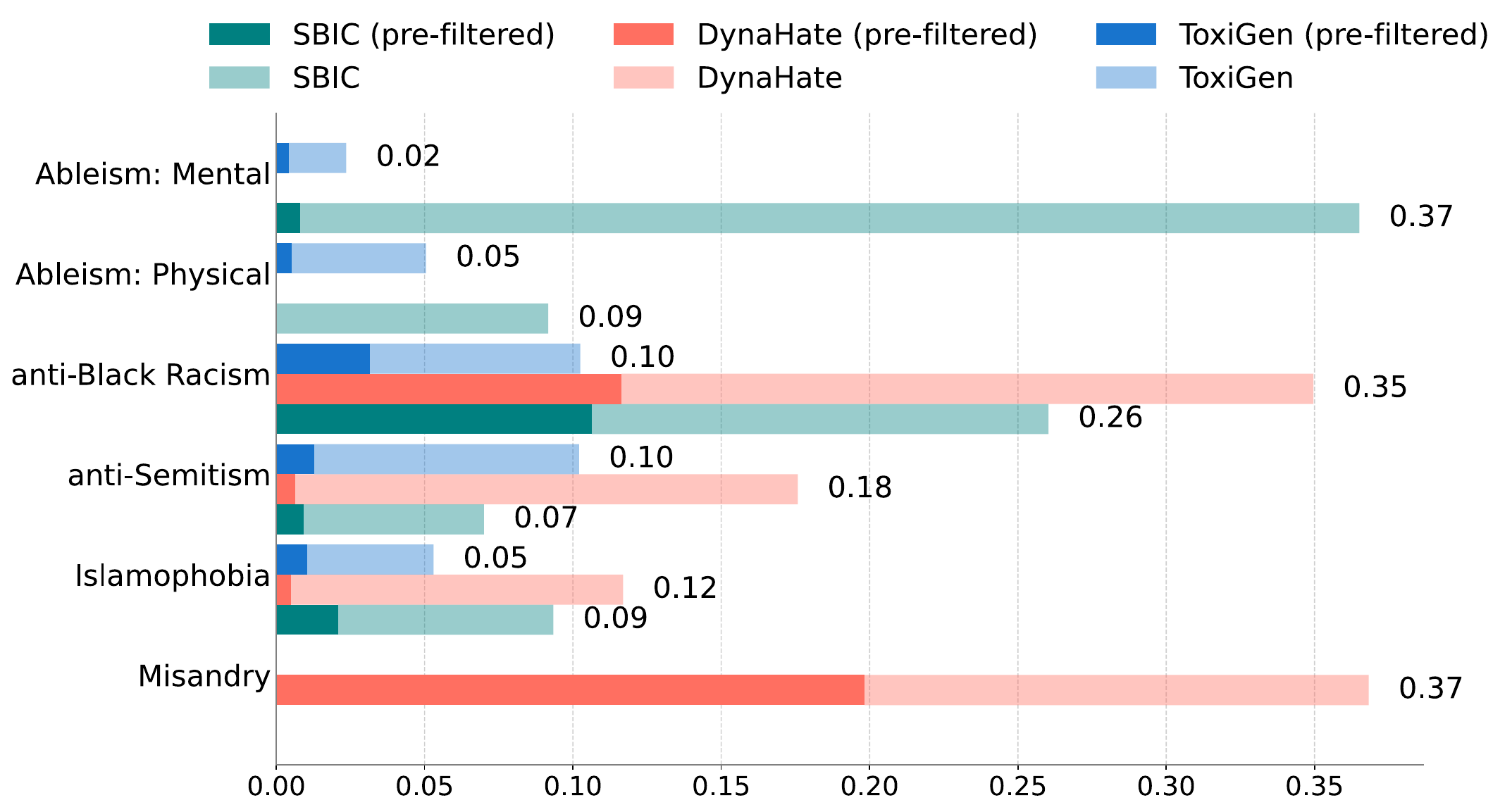}
       \caption{\textbf{\am's recall for subsets of hate examples directed towards various communities (All filters under the \emph{Discrimination \& Slurs} category turned on)}. The opaque portion of each bar indicates the fraction of messages that are pre-filtered. 
       }
       \label{fig:comm_recall}
\end{figure*}

 \textit{Filter-wise recall:} First, for each subset $\mathcal{D}_{c_i}$, we set $\mathcal{C}_A = \{c_i\}$ to measure the filter-wise recall (Table \ref{tab:exp_l2_relevant}). 
 We observe that the \textit{Misogyny} and \textit{Sex, Sexuality and Gender (SSG)} filters have the highest recall.
It is also important, however, to consider the pre-filtering rate. The Disability and Misogyny subsets have a very low pre-filtering rate on all datasets. This implies that most examples from these subsets went through the filter and the recall is a good representation of the filter-only performance. For the SSG filter, we observe the opposite --- the overall pre-filtering rate is $54.8$\%, and high across datasets. This implies that for detecting SSG related hate, \am relies on pre-filtering more heavily than it does for other kinds of hate speech. A similar inference cannot be made for the RER filter, however, because the pre-filtering rate for the RER filter on DynaHate is nearly half of what it is for SBIC. It is unclear whether the lack of diversity in the data itself causes higher pre-filtering in case of SBIC. We speculate that SBIC has more instances of racial slurs that appear in the pre-filtering blocklist than ToxiGen which mostly consists of implicit hate.

 \textit{Filter precision:} We repeat our experiment with $\mathcal{C}_A = \tilde{\mathcal{C}}$ for each $\mathcal{D}_{c_i}$. This allows us to measure the filter precision $P_{\mathcal{F}_i}$, the specificity of a filter. A naive filter that outputs 1 for every hateful text, will have a high recall yet is undesirable because it moderates examples beyond its criterion --- which may not be suitable when pedagogical or empowering utterances use n-grams that are frequently seen in hate speech. Measuring the specificity of filters is therefore as important as recall. The filter precision is shown in Figure \ref{fig:filter_precision}. We observe that the RER filter is the most precise, while Misogyny is the least.

\paragraph{Effects of moderation across target groups:} \label{subsec:level3}
In this analysis we study how \am performs in blocking hate related to specific target groups. To obtain hate data specific to different target groups, we use the target group labels in our dataset (mapping details in Appendix \ref{app:label_mapping}). While each subset may correspond to a filter (\textit{e.g.} anti-Black Racism $\rightarrow$ RER, Mental Ableism $\rightarrow$ Disability), there may also be instances where an example may be relevant to more than one filter. To address this in measuring target-group performance, we evaluate recall with $\mathcal{C}_A = \tilde{\mathcal{C}}$. By turning all filters on, we ensure that the recall under this setting is a fair measure of how well \am blocks hate directed towards each of these communities. The overall community-wise recall is shown in Figure \ref{fig:comm_recall}. We observe that hate speech directed towards Men, Black Folks and Mentally Disabled Folks is more effectively blocked by \am than for the other target groups.

\paragraph{Pre-filtering and Filter Level ($\alpha$):} We analyse the pre-filtered examples, and find the use of blocklists for pre-filtering causes disparate performance across target groups, perhaps due to bias in blocklist construction (see \cref{appsec: prefilter}). We also evaluate \am at different filter levels ($\alpha$) and find that for $\alpha > 0$, there is minimal gain ($+1.1\%$) in performance (\cref{appsec: ihc_filtration}).

\section{Case Studies: Implicit Hate, Context and Robustness}\label{sec:ablations}

This section explores three case studies that examine \am's versatility and robustness. 

\paragraph{Counterfactual Analysis:} 
\label{subsec: counterfactual-generation}

Unlike explicit hate speech, implicit hate is harder for systems to detect due to the usage of stereotypes, insinuations, or coded language --- which may evade traditional keyword-based filters. Therefore, detecting such content requires a deeper understanding of context, intent, and the subtle ways in which language can perpetuate harm. For this study, we select $110$ false negative samples from the SBIC dataset and manually review them to ensure that they are devoid of any slurs, while still remaining offensive. We then programmatically replace terms corresponding to demographic groups with offensive slurs associated with the same groups ( \emph{e.g.} replacing the phrase \textit{``Black People"} with the \emph{n}-word). When these counterfactual examples are passed to \am (with $\mathcal{C}_A = \tilde{\mathcal{C}}$), we observe $100$\% recall. Some examples from our counterfactual set are presented in Table \ref{tab:counterfac_examples}. This study highlights \am's heavy bias towards using slurs as an indication of hate and poor understanding of context which allows implicit hate to easily evade detection.

\paragraph{Policy Adherence Evaluation:}
\label{subsec: policy-discrepancy-section}
Twitch’s Community Guidelines explicitly recognize the importance of context in content moderation:

\begin{quotebox}
“At Twitch, we allow certain words or terms, which might otherwise violate our policy, to be used in an empowering way or as terms of endearment when such intent is clear.”
\end{quotebox}
To study \am's adherence to this policy, we manually select $20$ non-slur sensitive fragments from SBIC and prompt the GPT-4o model to generate statements that incorporate these fragments in a non-offensive manner including usage of the term in an educational or empowering context. The prompt used for generation and some examples are provided in \cref{app:counterfac_examples}. We test these examples (with $\mathcal{C}_A$ set to $\tilde{\mathcal{C}}$) with the filtering level varying between \textit{Some Filtering} ($\alpha = 2$) and\textit{ Maximum Filtering} ($\alpha=4$) for each filter. Under the former, \am flags $89.5$\% of examples while in the latter setting, it flags $98.5$\% examples. This behavior is in contrast to the aforementioned policy. While a streamer can manually configure \am to allow these sensitive fragments when a stream is likely to elicit their usage in non-offensive contexts, this could give a free pass to bad actors to perpetuate hate. This study further underscores the need for context-aware automated moderation.

\paragraph{Robustness:}\label{subsec: adversarial-experiment-section}We measure \am's robustness to minor, semantic-preserving perturbations  
of input text, given its seeming reliance on direct slurs. For this study, we prompt GPT-4o to make subtle changes (such as adding spaces or punctuation, altering spelling etc.) to $50$ manually-chosen sensitive fragments from the SBIC dataset. We employ 6 perturbation methods (listed in Table \ref{tab:adv_pert}). For each fragment we also generate one example that uses the fragment as is (Unperturbed). The prompt used for generation and the definitions of the perturbation methods are provided in \cref{appsec:data_Generation_prompts_for_adv_perturbation_experiment}. 
We observe that the recall drops from $100\%$ to $4\%$ on some of our perturbations. This indicates that malicious actors can easily evade \am. However, it is worth noting that \am was able to detect all perturbations (except the reverse one) of certain frequent terms such as the \emph{n}-word. This indicates some degree of robustness for highly sensitive terms.

\section{Discussion and Future Work}
\label{sec:discussion}

\am exhibits poor recall and F1-scores across datasets, compared to SoTA language models (\cref{fig:sota_comparison}), and even GPT-2~\cite{sap-etal-2020-social}. Our analysis of the false negatives and positives (\cref{sec:results} and \cref{appsec:fnfp}), along with the case studies (\cref{sec:ablations}) collectively explains how the poor performance of \am arises from a lack of contextual understanding of hate speech. Regardless, Twitch's effort in the development of \am is commendable for the flexibility it provides, with most other platforms lacking such tools. Our audit serves to highlight current gaps in \am's capabilities and we make good faith recommendations for Twitch to address these issues for greater user-safety.
There are several directions for future work, the most direct of which is to evaluate Twitch's Smart Detection feature for text, and expand the audit to audiovisual content. Considering other platforms and languages, particularly given the availability of multilingual hate speech data \cite{arora2023uli} is of critical importance. More nuanced audits for black-box systems may be possible by leveraging techniques such as model reconstruction attacks \cite{tramer2016stealing} for reverse engineering and transfer- and query- based black-box adversarial example generation \cite{demontis2019adversarial,bhagoji2018practical}
In conclusion, we hope our study serves as a blueprint for further audits of decision-making systems operating in complex socio-technical environments.

\section*{Limitations}\label{sec:limitations}

While we attempt to be comprehensive in our evaluation of hate speech moderation on Twitch, several key limitations remain. \emph{First}, we only consider a single platform in our study, and do not compare \am to any other deployed text moderation. As of the beginning of this study, we could not find any other suitable candidates. However, alternatives such as Mistral's Moderation tool \cite{mistralmoderation} have since emerged which would make for an interesting follow-up study, particularly in the context of other studies of open-source LLMs for moderation \cite{kumar2024watchlanguageinvestigatingcontent}. A concurrently published study \cite{Hartmann_2025} carries out an evaluation similar to ours for various moderation APIs on similar datasets. 
\emph{Second}, we largely investigate both hateful and benign messages in a non-conversational context, \emph{i.e.,} messages are passed one-by-one without any simulation of dialogue. More work is needed to curate datasets of dialogue that contain both hateful and benign content. For curation of such datasets, researchers may consider pursuing a DynaHate-style \cite{vidgen-etal-2021-learning} creation protocol, which is now feasible thanks to our framework that allows one to query black box systems like \am at scale. In addition, investigating in-group/out-group dynamics may uncover differential treatment and moderation of content based on group affiliations. In the case of \am, however, we speculate that our results would have remained unchanged as the policy violation case studies already demonstrate a lack of contextual awareness. \emph{Third}, our study is focused entirely on text in English. Considering other languages and modalities would be a critical direction for future work.

\section*{Ethical Considerations Statement}
\paragraph{Exposure to Offensive Content:}
In conducting this research, our team dealt with a significant amount of offensive content. All authors were aware of the nature of the work and consented to view such content. It is important to note that no individuals apart from the authors were exposed to this material given our use of siloed experimental streams. Moreover, the harmful content used in this work was sourced from open-source datasets.  

\paragraph{Legal Compliance:}
To conduct this research, we did post content that violates Twitch's terms of service. However, this action was taken within the legal boundaries established by the \emph{Sandvig v. Barr} case~\cite{sandvigvbarr}, which protects such research activity. 

\paragraph{Potential for Adverse Impact:}
We demonstrated how perturbations to offensive text circumvent Twitch's existing moderation systems, and there is a potential adverse impact wherein Twitch users might exploit these examples. However, these techniques and attacks are already well-documented and have been extensively explored in the literature concerning NLP classifiers. If a hateful actor is determined to spread hate on Twitch, it is unlikely that they are unaware of these rather unsophisticated perturbations.

\paragraph{Impact on Twitch Developers:}
We also recognize the unfortunate, unlikely possibility that our study inadvertently increased the burden for Twitch's internal moderation developers. 
Twitch has added measures to protect its users, and we do not wish to undermine these efforts. Instead, we intend to highlight existing failures while advocating for careful audits of deployed systems. We have contacted Twitch with our findings prior to the publication of this work, ensuring that they are aware of their system vulnerabilities and can take appropriate action.

\section*{Acknowledgments}

We thank anonymous reviewers for the constructive suggestions. 
YP is thankful for the Reliance Foundation Postgraduate Scholarship for supporting his research. GC is grateful to the Careers in Computer Science Mini-Internship Program at the University of Chicago for supporting her research. 
DP is additionally grateful for Adobe, Inc. for generously supporting his group's research. 
We note that the findings, conclusions and opinions presented in the paper are those of the authors and do not necessarily reflect the views of the sponsoring organizations or agencies.

\bibliography{ref}
\appendix
\clearpage
\appendix

This Appendix provides additional information corresponding to each section of the main paper as follows:
\begin{enumerate}
	\item \textbf{Platform Choice and Twitch Moderation Details} (\cref{appsec: twitch_details}): Provides additional details about Twitch Moderation options referenced in \cref{sec:twitch}.
	\item \textbf{Dataset Details} (\cref{app:data}): Detailed descriptions of the 4 datasets of hate speech content introduced in \cref{subsec: data}.
	\item \textbf{Experimental Setup Details} (\cref{appsec: setup}): Breakdown of the steps followed to send and receive messages from Twitch, expanding upon \cref{subsec: setup}.
	\item  \textbf{Further Results} (\cref{appsec: further_results}): Building on results in \cref{sec:results} with analysis of false negatives and false positives from \am, SBIC data threshold, different \am filter levels, prefiltering bias, and quality control analysis of filter-specific datasets.
	\item \textbf{Ablation Details} (\cref{appsec: further_ablations}): Information on the methods used to construct counterfactuals, policy adherence samples and perturbed examples in \cref{sec:ablations}.
	\item \textbf{Filter- and Community-Specific Subset Extraction} (\cref{app:label_mapping}): Dataset-wise breakdown of the procedure followed to obtain subsets for experiments in \cref{sec:results}.
\end{enumerate}

\begin{table*}[t]
 \centering
\begin{tabularx}{\textwidth}{cp{0.25\textwidth}C}
\toprule
\textbf{Platform} & \textbf{Closed environment} & \textbf{Moderation tools}\\ \midrule
Discord & \RaggedRight{Private server with \newline permissions for
everyone but admin disabled} & 
                    \RaggedRight{
                        Image hashing and ``ML powered tech" for child sexual abuse material; No native ML models for text, relies on keyword detection}\\
                        \midrule
Reddit & \RaggedRight{Via private subreddits} & 
                    \RaggedRight{ 
                         Needs plugins for ML models and subreddit specific rules for moderation; Well-investigated \cite{Kolla2024May, kumar2024watchlanguageinvestigatingcontent, Franco2023Dec}}\\ \midrule
Twitch & \RaggedRight{Untagged streams are private} & 
                    \RaggedRight{
                        Customisable moderation levels using native ML models; Also has keyword lists and Smart Detection to train model on the actions of human moderators}\\ \bottomrule
\end{tabularx}
\caption{\textbf{Comparison of candidate platforms.} We survey various platforms and choose Twitch primarily for its configurability and ML-based moderation tool.}
\label{tab: platform_survey}
\end{table*}

\section{Platform Choice and Twitch Moderation Details}\label{appsec: twitch_details}

As an extension of the Platform Exploration and Twitch Moderation section (\cref{sec:twitch}), we provide more details about the investigation guiding our platform choice, the Twitch interface, and its internal documentation. 

\subsection{Platform Survey}
In \cref{tab: platform_survey}, we provide details about the 3 platforms that we choose from in terms of their suitability for the audit. While both Reddit and Discord offer siloed environments for experimentation, neither provides native machine learning models for offensive text detection.

\subsection{Twitch Automated Moderation}

\begin{figure*}
    \centering
    \begin{subfigure}{0.48\textwidth}
        \centering
        \includegraphics[width=0.7\textwidth]{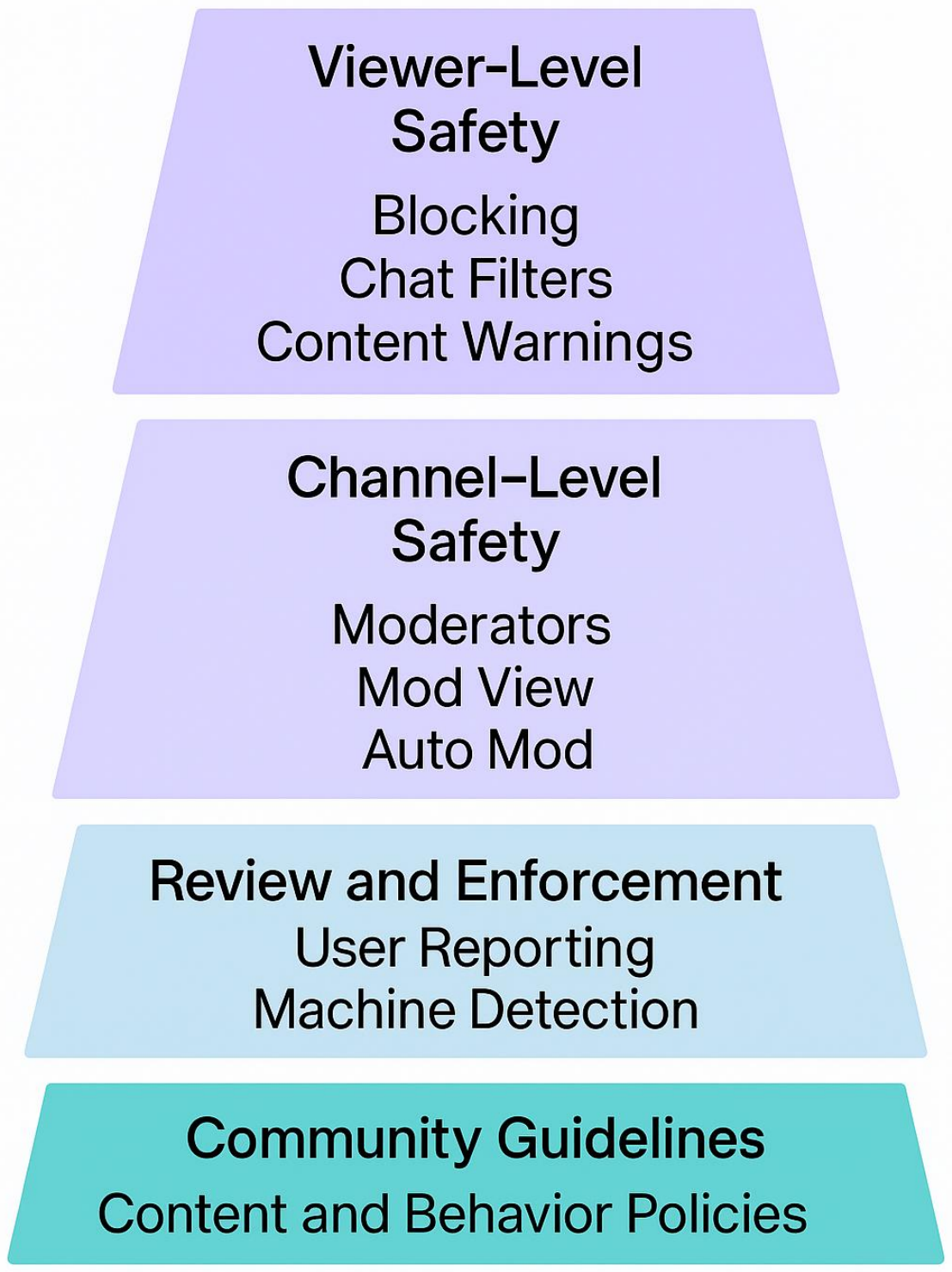}
        \caption{Twitch's moderation pyramid}
        \label{subfig:mode_pyramid}
    \end{subfigure}
    \hfill
    \begin{subfigure}{0.48\textwidth}
        \centering
        \includegraphics[width=1.1\textwidth]{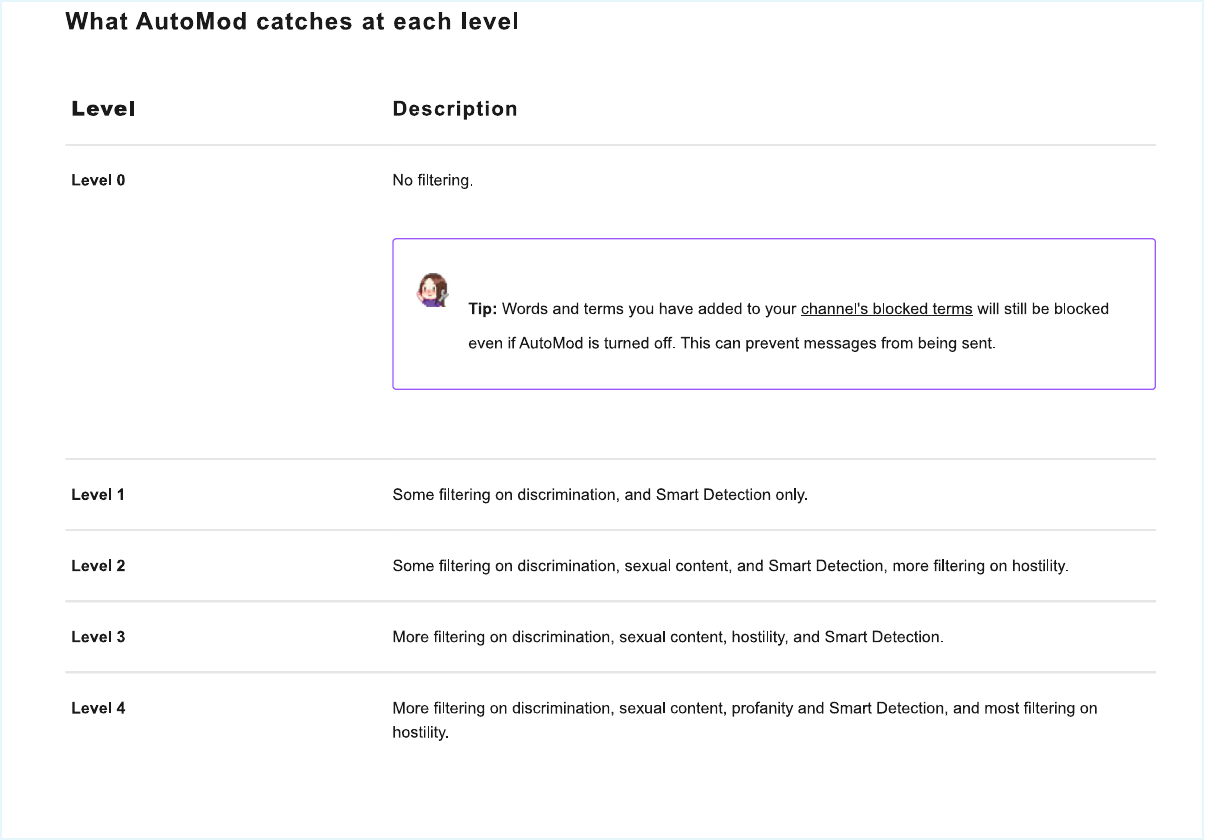}
        \caption{\am moderation level ($\alpha$) distinctions from Level 0 to 4}
        \label{subfig:automod_levels}
    \end{subfigure}

    \caption{\textbf{Official Twitch documentation of moderation tools:} (a) Different levels at which Twitch moderates content along with their respective methods of moderation (b) Twitch documentation on \am moderation levels (referred to as $\alpha$ in this work) from 0 to 4.}
    \label{fig:modDetails}
\end{figure*}

In \cref{subfig:mode_pyramid}, we show the Twitch Moderation Pyramid \cite{TwitchChatDoc}. As stated in \cref{sec:twitch}, we focus on viewer-to-streamer and streamer-to-streamer which are moderated via \am and Chat Filters. In \cref{subfig:automod_levels}, we show the various levels Twitch offers moderators to customize filtering strength.

\begin{table*}[t]
\begin{tabular}{p{0.25\textwidth}p{0.7\textwidth}}
\toprule
\textbf{Moderation Category} & \textbf{Explanation of Hate Speech Category}                                                                                 \\ 
\midrule
Sexual Content               & Words or phrases referring to sexual acts and/or anatomy.                                                                   \\ 
\midrule
Discrimination and Slurs     & Includes race, religion, gender-based discrimination. Hate speech falls under this category.                                 \\\\
\hspace*{5pt}I) Disability                   & Demonstrating hatred or prejudice based on perceived or actual mental or physical abilities.                                 \\
\hspace*{5pt}II) SSG   & Demonstrating hatred or prejudice based on sexual identity, sexual orientation, gender identity, or gender expression.       \\ 
\hspace*{5pt}III)Misogyny                     & Demonstrating hatred or prejudice against women, including sexual objectification.                                           \\ 
\hspace*{5pt}IV) RER & Demonstrating hatred or prejudice based on race, ethnicity, or religion.                                                     \\ 
\midrule
Hostility                    & Provocation and bullying, sexual harassment.                                                                                 \\ 
\midrule
Profanity                    & Expletives, curse words, and vulgarity. This filter especially helps those who wish to keep their community family-friendly. \\ 
\midrule
Smart \newline Detection              & Detects unwanted messages (including spam) based on moderation actions taken in your channel.                                \\ 
\midrule
\end{tabular}
\caption{\textbf{\am Content Categories.} The definition of various categories and sub-categories that \am moderates.}\label{tab: automod_categories}
\end{table*}

\cref{tab: automod_categories} describes Twitch's Official Content Categories and their differentiating definitions as of $04$ July, $2024$. The audit's investigation focuses on the effectiveness of \am's overall performance with respect to the Discrimination and Slurs content category. More specifically, the datasets used provide ground truth relating to the subsections of Discrimination and Slurs as seen in \cref{tab: automod_categories}, Disability, SSG, Misogyny, and RER.

\begin{figure*}[t]
    \centering
    \includegraphics[width=2\columnwidth]{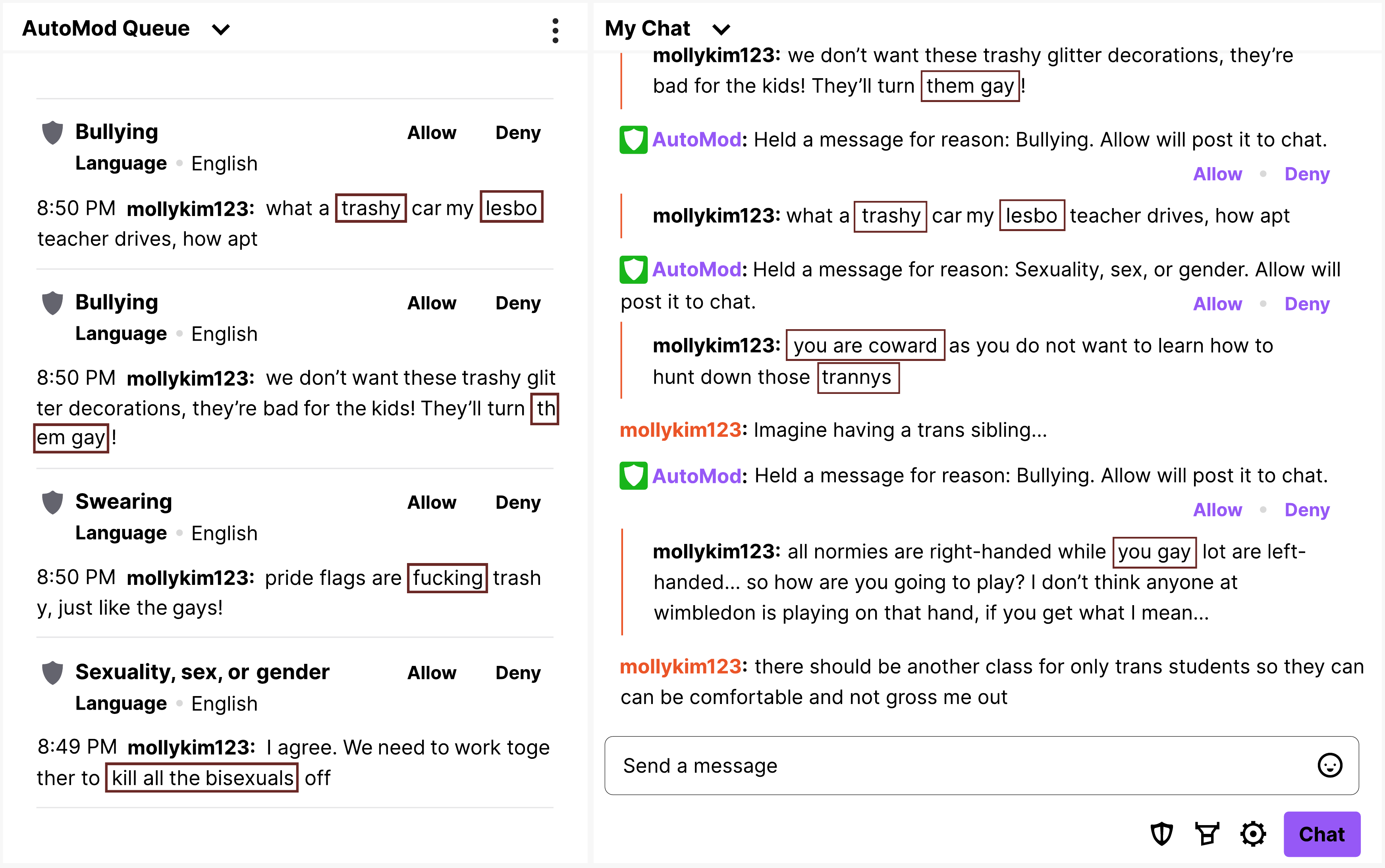}
    \caption{\textbf{Moderation view as seen from a streamer's account.} (\textbf{Note}: The original screenshot, which was in dark mode UI was manually adapted to light mode for print clarity; minor visual differences from the actual light mode may exist.)}
    \label{fig:moderatorView}
\end{figure*}

The dashboard is customizable allowing the streamer to view in real-time their stream's \am queue and to manually approve or deny detected messages as shown in \cref{fig:moderatorView}. The \am queue also displays the associated content category so as to inform the streamer for \am's reasoning behind moderation. Highlighted segments also help the user focus on the problematic fragments of moderated messages.

\begin{figure*}[t]
    \centering
    \includegraphics[width=\linewidth]{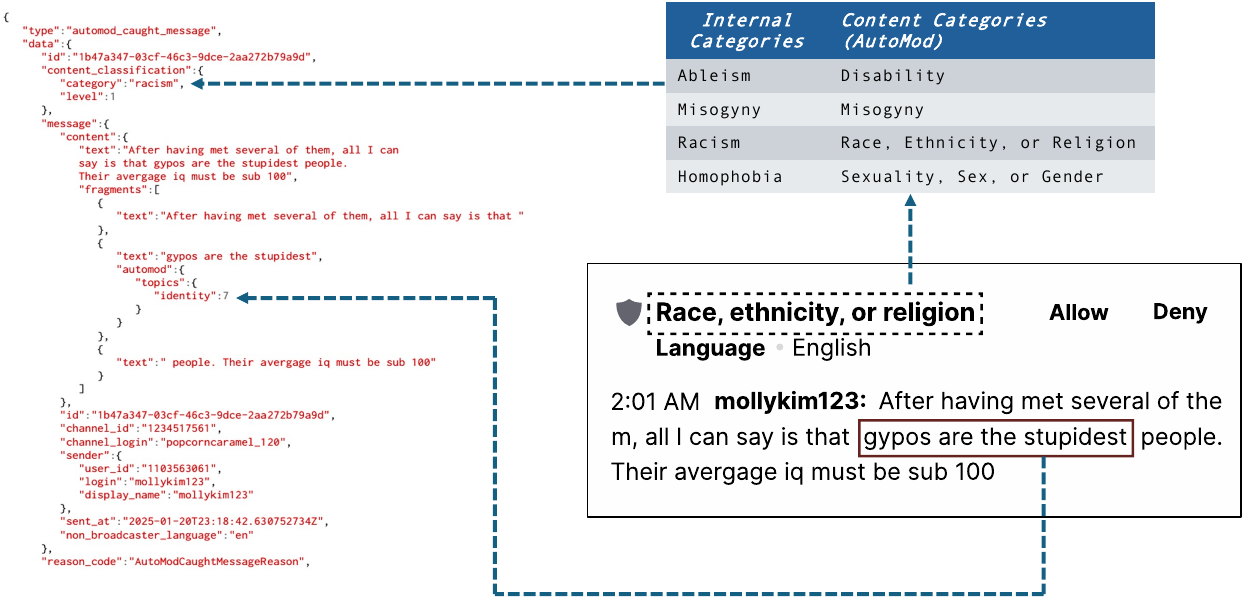}
    \caption{\textbf{Correlating the PubSub output with moderation decisions}: (left) Sample JSON output received from PubSub; (top right) Mapping of JSON output to \am's content categories (described in \cref{tab: automod_categories}); (bottom right) Mapping the \texttt{automodqueue} (visible to the streamer via the UI) to the JSON obtained from PubSub.}
    \label{fig:pubsub_result}
\end{figure*}

In evaluating the overall performance of \am on moderating various categories of hateful content, mere binary evaluations were deemed primary but not comprehensive to effectively determine performance in terms of accuracy and precision. Hence, \am's moderation reasons with respect to Twitch's pre-defined categories of moderation provided more insight. To programmatically access the streamer's \am queue, Twitch Pubsub event subscriptions were used to grab API results whenever \am logged a message in the queue for moderation. However, we found that \am's internal categories did not match Twitch Documentation and Policies on Content Moderation across the platform. \cref{fig:pubsub_result} depicts a snapshot of Twitch API result from \verb|automodqueue| event subscription detailing the message and its metadata. This metadata include \verb|topics| and \verb|category| of the moderated message from which we map the categories semantically.

\section{Dataset Details} \label{app:data}
All datasets used are open-source and permitted for research. All of our data is in English.

\subsection{Dynahate}
The Dynahate dataset~\cite{vidgen-etal-2021-learning} contains approximately 41,000 entries, each labeled as either 0 (Not hateful) or 1 (Hateful) and is generated and labelled by trained annotators over four rounds of dynamic data creation. The dataset includes 54\% of hateful examples.  In our work, we use three columns from the Dynahate dataset: \textit{text}, \textit{label}, and \textit{target}. The `target' column specifies the community towards which the hate is directed, making it particularly useful for dividing the data into filter and community-specific subsets.

\subsection{SBIC}

The Social Bias Inference Corpus (SBIC)~\cite{sap-etal-2020-social} is a large-scale dataset with over 150k annotations across 34,000 social media posts, capturing nuanced biases and stereotypes in online language. Its focus on offensive content, implied stereotypes, and target demographics makes it particularly useful for auditing Twitch's moderation system, especially in areas related to group-based bias and contextual offensiveness. We sample 20k examples from the SBIC training data for our work. Since SBIC provides an average annotator offensiveness score, we set a threshold of 1 (all annotators agreeing on the offensiveness of an example) to evaluate the overall effectiveness of \am, resulting in a hate-to-non-hate ratio of 1:3. For further division into hateful subsets related to specific filters, we use a threshold of 0.5, yielding a total of 8,748 examples. The SBIC dataset contains numerous features, but for our analysis, we utilize the following: \textit{post}, \textit{targetMinority}, \textit{targetCategory}, \textit{offensiveYN} (indicating whether the example is offensive, non-offensive, or ambiguous), and annotator-related columns for aggregating the offensive score.

\subsection{ToxiGen}
The ToxiGen dataset~\cite{hartvigsen-etal-2022-toxigen} is a large-scale dataset containing approximately 274k toxic and benign statements about 13 minority groups. This dataset is particularly useful for our work as it consists of synthetically generated implicit hate speech, allowing us to evaluate the effectiveness of filters on implicit hate content. The authors use the GPT-3 model to generate implicit hate examples by providing human-curated toxic and non-toxic examples as prompts. For detailed information on the data generation methods, we refer the reader to the ToxiGen paper. ToxiGen includes various features, but we use the following in our work: \textit{prompt}, \textit{generation} (text generated by the model—examples we use for auditing), \textit{prompt label}, and \textit{roberta prediction} for the generated sentence. For our experiment, we randomly sample 20k hateful examples from the dataset, selecting entries with toxic prompts and a \textit{roberta\_prediction} score between 0.8 and 1.0. Similarly, we sample 20k non-hateful examples with non-toxic prompts and a \textit{roberta\_prediction} score between 0 and 0.2.

\subsection{Implicit Hate Corpus(IHC)}
The Implicit Hate Corpus (IHC)~\cite{elsherief-etal-2021-latent} is a dataset containing approximately 6,000 examples of implicit hate. For our experiments, we additionally include a random sample of 6,000 benign examples from the dataset. It includes both the target and implied meaning of hateful statements, collected from hate communities and their followers on Twitter. In our work, we use this dataset to evaluate the effectiveness of all discrimination filters. Since the dataset is not large enough to be divided into filter-specific subsets, we only remove examples that are not aligned with discrimination. From this dataset, we utilize the \textit{post} and \textit{target} columns.

\section{Experimental Setup Details}\label{appsec: setup}
Here we describe in detail the setup and components needed to build the experimental pipeline (\cref{fig: experimental_design}) that was briefly described in \cref{subsec: setup}. The critical steps for implementation include: $1)$ bot creation to send messages, including adherence to security measures, registration, and scope handles, $2)$ \am event subscriptions for sorting messages into moderated and unmoderated, and $3)$ results handling.

\subsection{Sending Messages Programatically} The following steps are needed to \emph{send} messages programatically at scale:

\textbf{Step 1: Create a chatbot} In order to handle the creation of chatbots, Twitch requires a two-factor authenticated phone number and email address for each account. The bots created for implementation utilize free online phone numbers for 2FA, Burner phone numbers (Verified Phone Numbers only), and personal phone numbers. The different bots also require an application registration in the associated verified-account's Twitch Developer Portal. This process requires the specification of a unique chat application name, OAuth Redirect URL specification: \textit{https://localhost:3000}, and the bot's functionality: \textsf{Chat Bot}. This application enables Twitch Developers to extract application-specific Client-ID, and Client-Secrets; 

\textbf{Step 2: Authenticate and register the chatbot} Twitch enables authentication under OAuth$2.0$ using Twitch OIDC Authorization Code Grant Flow in order to grant the application specific access to Twitch HTTPS resources. This access, for example, is specified as \texttt{chat:read},\texttt{chat:write} access for IRC Chatbots. Once the Authorization code from HTTPs POST requests to \texttt{https://id.twitch.tv/oauth2/token} is received, the app access token and refresh tokens are retreived and used for API calls. As each application's access token used to connect to Twitch Chat and AutoMod queues, we automatically handle the renewal of these authorized tokens using the application's refresh token and reconnect accordingly when its expiration time has reached; 

\textbf{Step 3: Send messages at scale with appropriate timing configuration} We create one instance of the experimental pipeline by including a messenger bot, channel bot, and Pubsub bot (three-bot configuration). The messenger bot which sends up to $5$ messages with a $4$ second wait between each message. At every iteration, we pause for $3.5$ seconds due to Twitch's Chat Rate Limits \cite{TwitchChatDoc} allowing for less than $20$ messages per $30$ seconds for normal chat bot accounts. This ensures that each message is sent and processed into the stream chat without duplication or misses due to connection delays. We conduct iterative tests and conclude that this configuration of pauses and message counts allows for prolonged experiment run times due to the large number of messages from our four datasets.

\subsection{Receiving moderated and unmoderated messages} To receive and sort the messages we need to create additional bots and parse their outputs. This connection is facilitated by the use of \texttt{tmi.js} JavaScript package that creates a connection to the Twitch IRC server using tmi.js servers \cite{tmiJsDOC}. Twitch IRC presents a reduced-functionality RFC1459 and IRCv3 Message Tag specification \cite{TwitchChatDoc} for parsing messages to and from a specified Twitch channel. This extraction of non-moderated messages is handled by the Receiver Bot. The Pubsub bot uses PubSub event subscription to extract messages from the \am queue in real time which are accessible from the channel's creator dashboard. Using twitchAPI.js, we enable a websocket connection to Twitch machines providing Pubsub services from which registered chat bots listen for \am queue items and receive message metadata in a websocket frame.

\subsection{Parsing results}
The json received from Pubsub provides information about the moderated fragment, and an internal label for moderated content such as \texttt{Ableism, Misogyny, Racism, and Homophobia} (\cref{fig:pubsub_result}). While these categories are not detailed in \am documentation, we infer an injective relationship between the content categories in \am documentation and internal categories 
from the API call. With these labels, we are able to investigate \am's stated reasons for moderation and conduct further category-specific analysis in~\cref{subsec: level2}.

\subsection{Challenges} Despite being under the rate limit, we suspect that the increase in fraudulent activity across Twitch as reported in 2024's Transparency Report \cite{twitchtransparency24} has led to the inaccurate but frequent permanent banning of our chat bots. As a result, this lead to the creation of at least 30 different accounts. After conducting several initial experiments and analyzing the results, we discovered that certain messages never appeared in the receiver bot's inbox and, consequently, did not trigger any Pub/Sub events. To investigate further, we manually passed some of these messages and found that Twitch had prefiltered them, preventing them from being sent.To address this, we enhanced our code to log all messages—both those moderated by AutoMod and those that were not—into a CSV file. By comparing this log against the original input data and filtering out the missing messages, we successfully identified the set of prefiltered messages.

\section{Further Results}\label{appsec: further_results}
All our experiments with language models were carried out on a workstation with $6$ A6000 GPUs.

\subsection{Evaluating the Chat Filter}\label{app:chat_filter}
\begin{figure}[t]
    \centering
    \setlength{\fboxsep}{-2pt}%
    \setlength{\fboxrule}{0.2pt}%
    \begin{subfigure}{\columnwidth}
        \centering
        \fbox{\includegraphics[width=\textwidth]{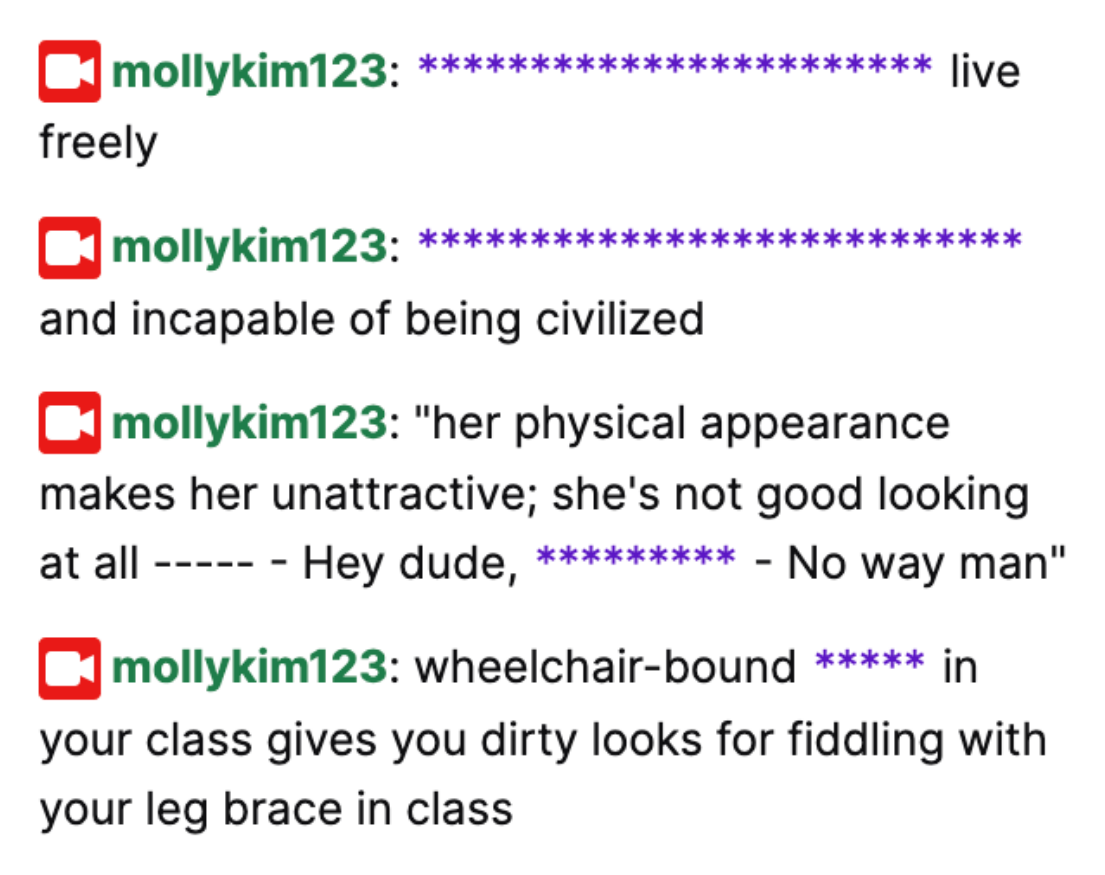}}
        \caption{Twitch Simulation: User with chat filters turned on for Disability, Misogyny, RER, and SSG excluding profanity\\}
        \label{subfig:chatFilter_mod}
    \end{subfigure}

    \vfill

    \begin{subfigure}{\columnwidth}
        \centering
        \fbox{\includegraphics[width=\textwidth]{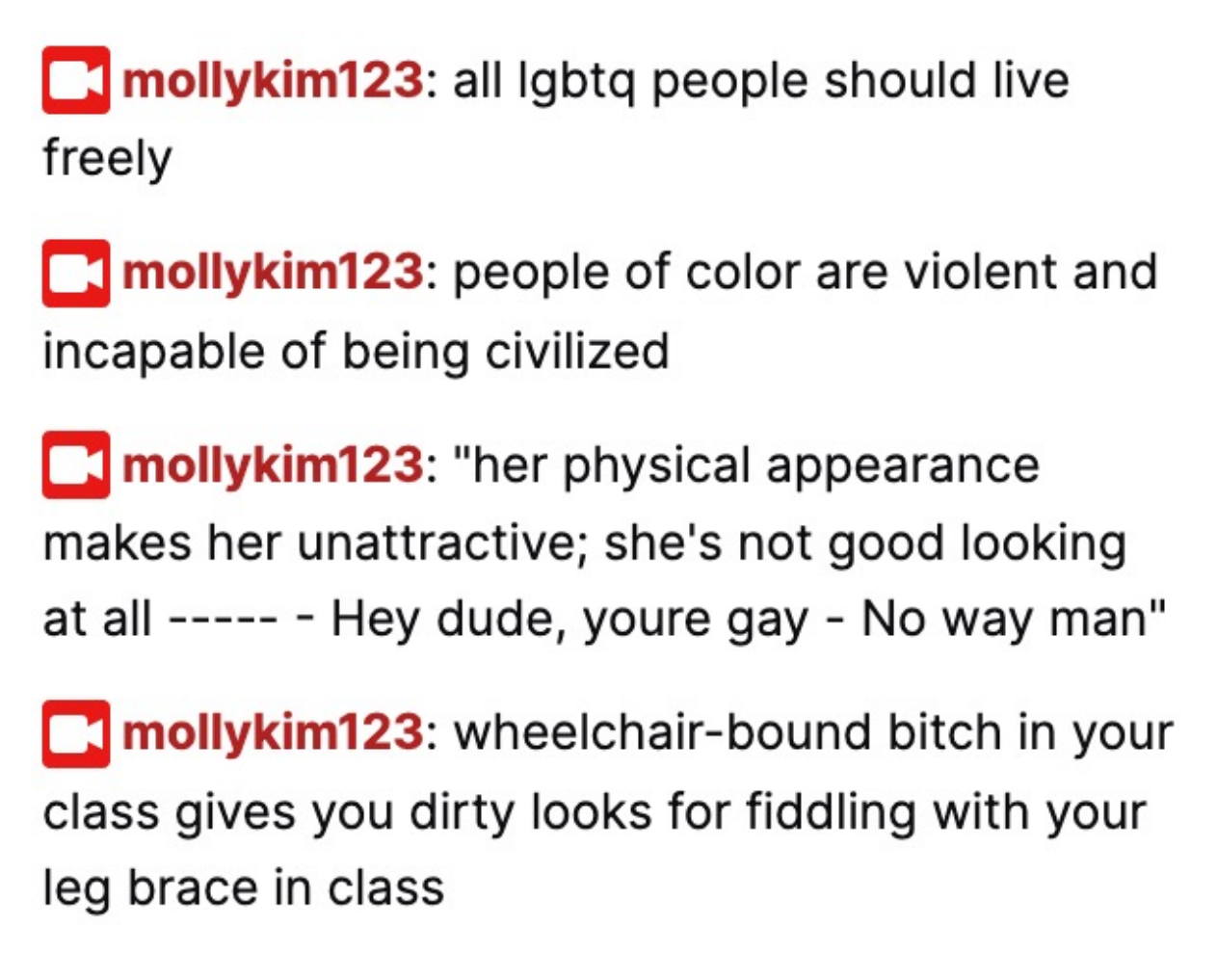}}
        \caption{Twitch Simulation: User without chat filters on}
        \label{subfig:chatFilter_notMod}
    \end{subfigure}

    \caption{\textbf{Example of user-chat with and without chat filters.} We observe that the chat filter behaves similar to \am whereby false positives are prevalent and fails our target group analysis standards as seen in (a). Original Messages are provided in (b)}
    \label{fig:chatFilter}
\end{figure}

The viewer-controlled chat filter can only act at the granularity of the broad \emph{Discrimination and Slurs} category, and does not trigger events for the Pubsub bot we use to detect moderation events which makes experiments at scale challenging as manual inspection is required to identify moderation, indicated by a ``***'' (\textit{star pattern}).
To audit the alignment between \am and the Chat Filter, we randomly sample $200$ examples from SBIC consisting of equally many moderated and non-moderated examples and pass them through the chat filter. We observe that the Chat Filter is unable to moderate any of the non-moderated examples (\cref{fig:chatFilter}). We infer that the Chat Filter behaves similarly to \am at $\alpha=4$. 

\subsection{Filter Precision} \label{app:filter_precision}
The filter precision for filter $\mathcal{F}_i$, denoted by $P_{\mathcal{F}_i}$ is defined as:
$$P_{\mathcal{F}_i} = \frac{\sum\limits_{x \sim \mathcal{D}_{c_i}} \mathds{1}(\mathcal{F}_i(x) = 1)}{\sum\limits_{j \in \widetilde{\mathcal{S}_c}}\left(\sum\limits_{x \sim \mathcal{D}_{c_j}} \mathds{1}(\mathcal{F}_i(x) = 1)\right)}$$
 
\begin{figure}[H]
    \centering
    \includegraphics[width=\columnwidth]{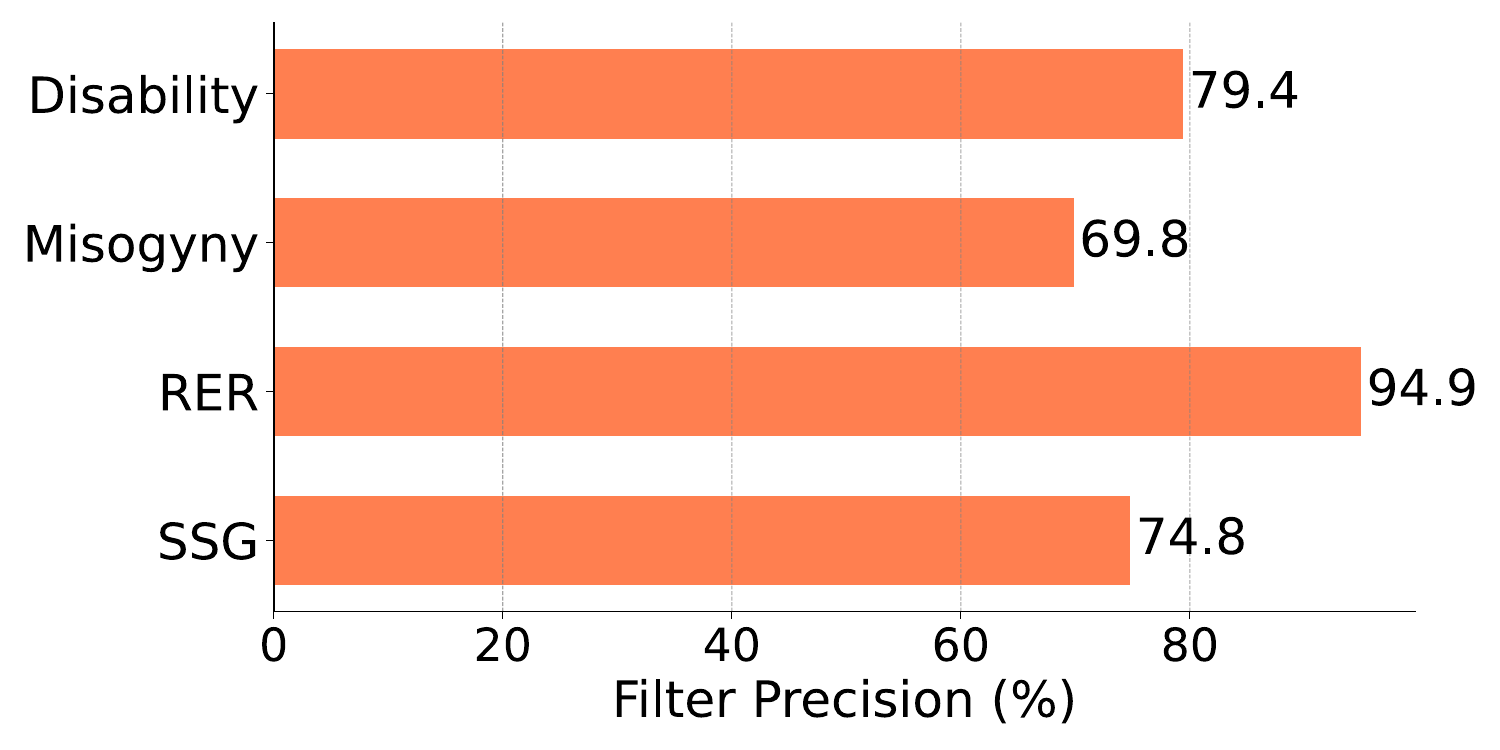}
    \caption{\textbf{The precision of filters in \am}. Precision measures the specificity of a filter. We find that the RER (Race, Ethnicity and Religion) filter is the most specific, while the Misogyny filter is the least specific.}
    
    \label{fig:filter_precision}
\end{figure}

It is the number of examples adhering to criterion $c_i$ (i.e., from $\mathcal{D}_{c_i}$) that were moderated by filter $\mathcal{F}_i$ upon the number of examples across all subsets that were moderated by filter $\mathcal{F}_i$. The precision of each filter is shown in Figure \ref{fig:filter_precision}. 

\subsection{False Negative and False Positive Analysis}\label{appsec:fnfp}

\subsubsection{False Negative Analysis}\label{appsec:false_negative_analysis}

\paragraph{Setup} For this experiment, we use three of our datasets --- SBIC, DynaHate and ToxiGen. We randomly choose a dev set of 10,000 examples from these three datasets and use the rest as test data. We use the \texttt{dev set} to tune our prompt and develop our answer extractor. We prompt the \texttt{llama-3.3-70B-instruct} model \cite{grattafiori2024llama3herdmodels} to identify if the provided text contains swear word(s) and record the model outputs for the test data. During both \texttt{dev} and \texttt{test} phases, we only use the False Negatives. 

\paragraph{Prompt} The prompt used for this experiment was a zero-shot prompt that instructed the model to identify the presence of swear word(s) in the provided text and return a binary label. The exact prompt is shown in Figure \ref{fig:sw_detection_prompt}
\begin{figure}
\centering
\begin{promptbox}[Swear Word Detection Prompt]
\textbf{System Prompt:} "You will be given a text. Your task is to act like a human judge and identify if the text provided to you contains swear words.  
Your response should be a dictionary with a single key - \texttt{"has\_swear\_words"} which corresponds to whether the provided text has swear words.  
You should only respond in the format as described below. DO NOT RETURN ANYTHING ELSE. START YOUR RESPONSE WITH `\texttt{\{}'.  

[response format]:  

\texttt{\{}  

\quad \texttt{``has\_swear\_words"}: "True" if the provided text has swear words else "False" 

\texttt{\}}"

\textbf{User Prompt:} "The provided text is:

[text]: \texttt{\{text\}}"
\end{promptbox}
\caption{\textbf{Prompt used for swear word detection.}}
\label{fig:sw_detection_prompt}
\end{figure}

\subsubsection{False Positives}
\am exhibits a low False Positive Rate (FPR) overall, and most of the false positives occur in the DynaHate dataset --- on which \am has a relatively higher FPR. Similar to the False Negative analysis, we hypothesize this is due to a high number of swear words in DynaHate and \am's overreliance on profanity as a signal for hate. We perform an experiment similar to the False Negative Analysis to detect the presence of swear words on DynaHate's false positive set. We find that nearly $73\%$ of the false positives from DynaHate contain profanity.
\paragraph{Setup} The setup remains the same as \cref{appsec:false_negative_analysis}, except that we use only those examples from the test set which are false positives from the DynaHate dataset.

\subsection{Performance Metrics at different values of SBIC threshold}\label{appsec: sbic_threshold}

\begin{figure}[h]
    \centering
    \includegraphics[width=\columnwidth]{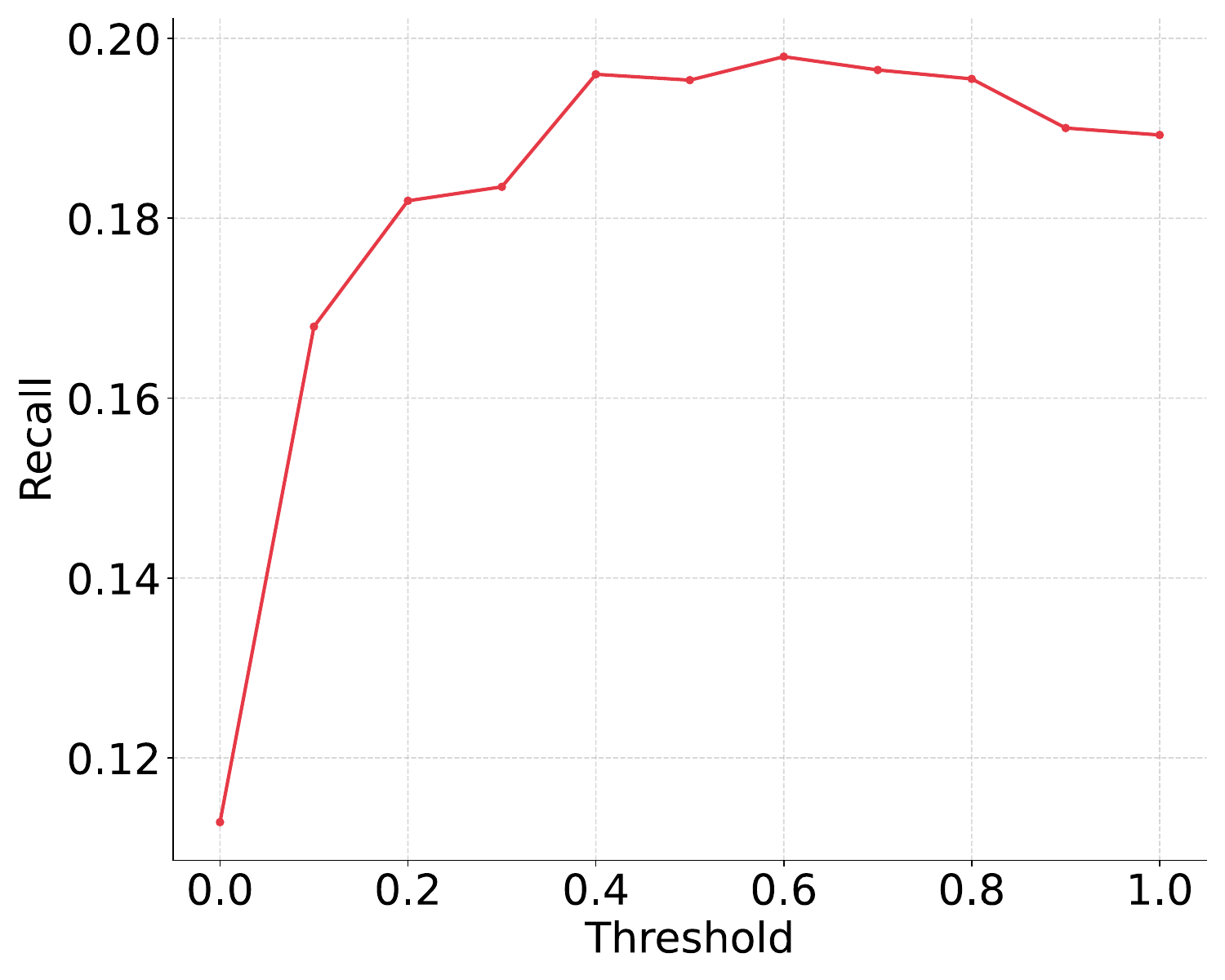}
    \caption{\textbf{\am's recall (with $\mathcal{C}_A=\tilde{\mathcal{C}}$) at different offensiveness score thresholds used for obtaining SBIC ground truth labels}. We see that recall reduces when more nuanced examples (lower threshold) of hate are included.}
    \label{fig:recall_threshold}
\end{figure}

The SBIC dataset provides annotator agreement scores for the offensiveness of each sentence, which we used to classify examples: sentences with agreement scores below threshold were categorized as non-hateful, while those with scores of threshold or greater were classified as hateful. See Figure ~\ref{fig:recall_threshold} for Moderation Rate(Recall) on different values of threshold for SBIC.

\subsection{Performance Metrics at different filter levels for IHC Dataset}\label{appsec: ihc_filtration}

Twitch offers five levels of filtration, with each filter adjustable from 
\textbf{No Filter} ($\alpha=0$) to \textbf{Maximum Filter} ($\alpha=4$) (See Figure \ref{subfig:automod_levels}). 
In our experiment, we apply the IHC dataset to all five filtering levels and measure recall for each, as shown in Table~\ref{tab:IHC_recall_at_levels}. 
The variation across the different levels is minimal, except for $\alpha=0$, where \am does not perform any moderation, and only the prefiltered text undergoes moderation.

\begin{table}[H]
\centering
\begin{tabular}{lr}
\toprule
\textbf{Filter Level} & \textbf{Recall (\%)} \\ 
\midrule
No filter             & 0.18                 \\
Less filter           & 5.31                 \\
Some filter           & 6.31                 \\
More filter           & 6.40                 \\
Maximum filter        & 6.42                 \\ 
\midrule
\end{tabular}
\caption{\textbf{\am's recall at different filter levels ($\alpha$)}. Recall values were calculated on the IHC dataset on for $\alpha=0$ to $\alpha=4$.}
\label{tab:IHC_recall_at_levels}

\end{table}

\subsection{Experimental Setup for computing performance metrics of SoTA language models}\label{app:benchmarking}
We choose 7 different language models with parameters ranging from 2B to 70B to compare with \am. We use the instruct variants of the models wherever available. We choose a random sample of $10,000$ examples as a test set. We ensure that the test set is balanced between the two classes. The numbers shown for \am in Figure \ref{fig:sota_comparison} are also correspond to this test set. We prompt the language models with Twitch's Community guidelines and ask to identify if the provided comment violates the guidelines.\footnote{For the prompt, we exclude part(s) of the Community Guidelines that are not relevant for text based moderation (e.g., guidelines for images/videos)} We use a zero-shot prompt for our experiment. We set temperature = 0 for reproducibility and use a \texttt{top\_p} value of 1. For models which support system prompts, we add the instruction and community guidelines in the system prompt and the comment to be labeled in the user prompt. For other models, we add all text in the user prompt itself. The prompt is shown in Figure \ref{fig:sota-prompt} and the performance metrics of the different models alongside \am are shown in Figure \ref{fig:sota_comparison}. To avoid redundancy, we only show the prompt in the System + User format. 

\begin{figure*}[htbp]
\centering
\begin{promptbox}[Moderation Prompt]
\textbf{System Prompt:} "You will be given a text. Your task is to act like an online human moderator on Twitch and judge if the provided comment conforms to the community guidelines of the platform (Twitch):

\textit{\#\# Community Guidelines:}

Twitch does not permit behavior that is motivated by hatred, prejudice or intolerance, including behavior that promotes or encourages discrimination, denigration, harassment, or violence based on the following protected characteristics: race, ethnicity, color, caste, national origin, immigration status, religion, sex, gender, gender identity, sexual orientation, disability, serious medical condition, and veteran status. We also provide certain protections for age, which are expressly noted in the examples.

We define `protected groups' as a subset of the population with a shared protected characteristic. Every Twitch user falls into multiple of these protected groups. Twitch affords every user globally equal protections under this policy.

For example, regardless of your intent, you may not:
\begin{enumerate}
    \item Promote, glorify, threaten, or advocate violence, physical harm, or death against individual(s) or groups on the basis of a protected characteristic, including age.
    \item Use hateful slurs, either untargeted or directed towards another individual. We allow certain words or terms, which might otherwise violate our policy, to be used in an empowering way or as terms of endearment when such intent is clear. We also make exceptions for slurs in music—and singing along to music—as long as the song itself is not hateful and the slurs are not combined with other discriminating or denigrating content. 
    \item Post a comment that dehumanizes or perpetuates negative stereotypes and/or memes.
    \item Post a comment that expresses inferiority based on a protected characteristic, for example, statements related to physical, mental, and moral deficiencies.
    \item Call for subjugation, segregation or exclusion, including political, economic, and social exclusion/segregation, based on a protected characteristic, including age. We do, however, allow discussions on certain topics such as immigration policy, voting rights for non-citizens, and professional sports participation as long as the content is not directly denigrating based on a protected characteristic.
    \item Encourage or support the political or economic dominance of any race, ethnicity, or religious group, including support for white supremacist/nationalist ideologies. This does not include support for self-determination movements.
    \item Expressions of contempt, hatred, or disgust based on a protected characteristic.
    \item Mock the event/victims or deny the occurrence of well-documented hate crimes, or deny the existence of documented acts of mass murder/genocide against a protected group.
    \item Make unfounded claims assigning blame to a protected group, or that otherwise intends to incite fear about a protected group as it relates to health and safety.
    \item Encourage the use of or generally endorsing sexual orientation conversion therapy.
    \item Support, promote, or be a member of a hate group, including sharing hate group propaganda materials.
\end{enumerate}

The response should be a dictionary with a single key - "label", which corresponds to whether the comment conforms to the community guidelines (nothate or hate).

You should only respond in the format as described below. DO NOT RETURN ANYTHING ELSE. START YOUR RESPONSE WITH '\{'. 

[response format]: 

\{

\quad    "label": `hate' if the comment violates community guidelines, else `nothate'

\}"

\textbf{User Prompt:} "The following is the given comment:

[text]: \texttt{\{comment\}}"

\end{promptbox}
\caption{\textbf{Moderation prompt for SoTA language models based on Twitch's community guidelines.}}
  \label{fig:sota-prompt}
\end{figure*}

\subsection{Prefiltering Bias}\label{appsec: prefilter}
\begin{figure}[H]
    \centering
    \includegraphics[width=\columnwidth]{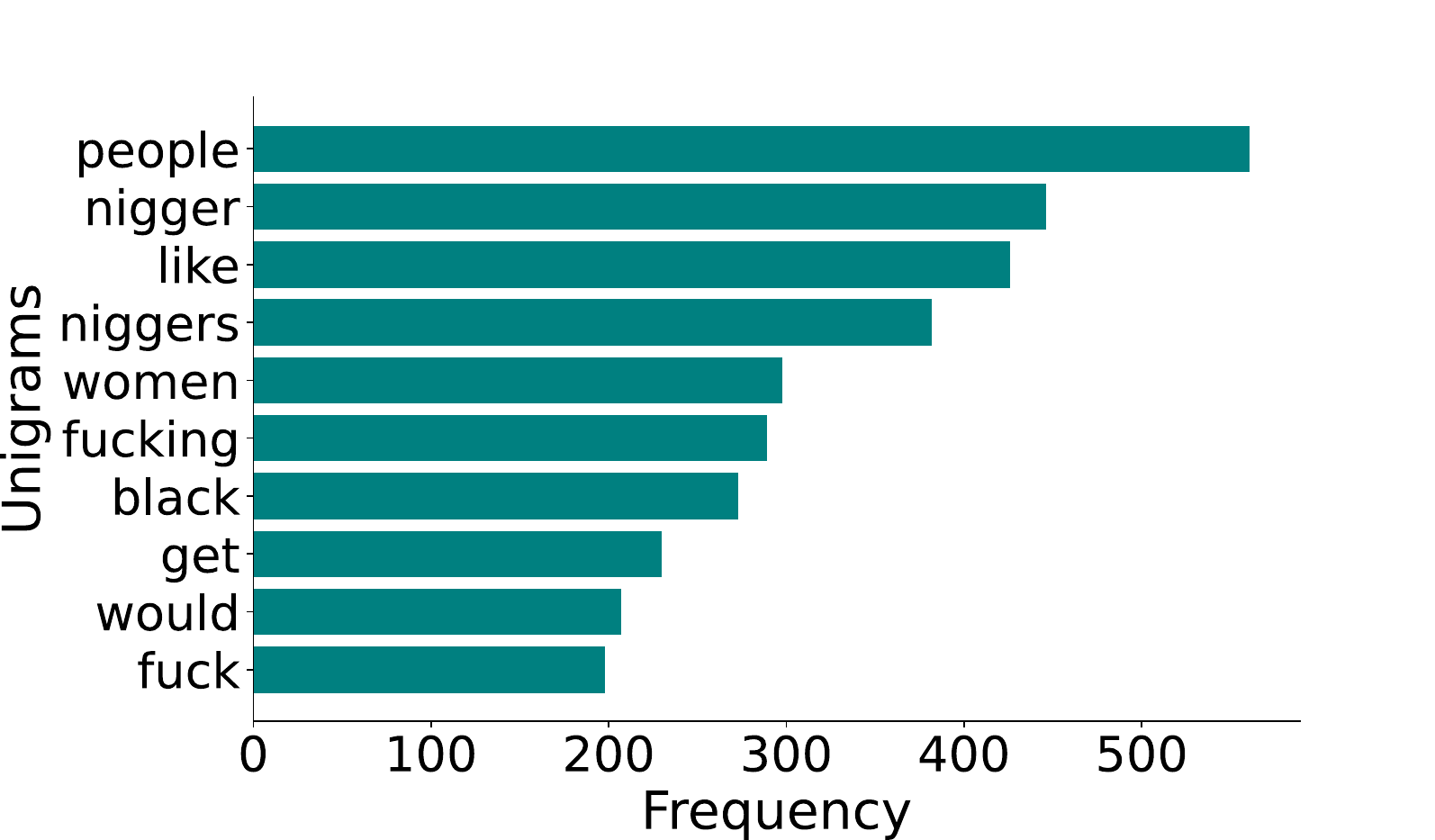}
    \caption{\textbf{Top 10 pre-filtered unigrams based on frequency.} Stop words were excluded while ranking unigrams according to frequency. We observe that both variants of the n-word combined together appear close to 800 times. }
    \label{fig:prefilter_bias}
\end{figure}

The \textit{n-word} is a highly offensive racial slur and is considered inappropriate in nearly all contexts---barring anti-racist uses in, for example, art. The n-word's infamy may push system designers to pre-filter the term as well as its derivative forms. We observe a high frequency of the \textit{n-word} in the pre-filtered examples as shown in Figure~\ref{fig:prefilter_bias}. While such system designs are done in good faith, they may induce biases in the system when the channel level moderation algorithms are not sophisticated enough and the blocklist used for pre-filtering is not exhaustive.\footnote{Here, by `blocklist' we refer to the implicitly/explicitly defined list of words that the pre-filtering algorithm filters (Figure~\ref{fig:prefilter_bias}).} We hypothesize that due to the presence of the \textit{n-word}, the system will perform much better at blocking hate directed towards Black Folks than it would for the other communities. To test this hypothesis, we set $\mathcal{C}_A = \tilde{\mathcal{C}}\setminus \{c_{\text{RER}}\}$ and record the outputs. We observe that on both SBIC and DynaHate combined, $37.4\%$ of the recall on Black-related hate speech is due to prefiltering. For the Jewish and Muslim categories, this number is much lower ($6.1\%$ and $8.6\%$ respectively).

\subsection{Quality Control analysis for data subsets} \label{app:data_quality_control}

\begin{figure*}
   \centering
  \includegraphics[width=\textwidth, height=0.5\textwidth]{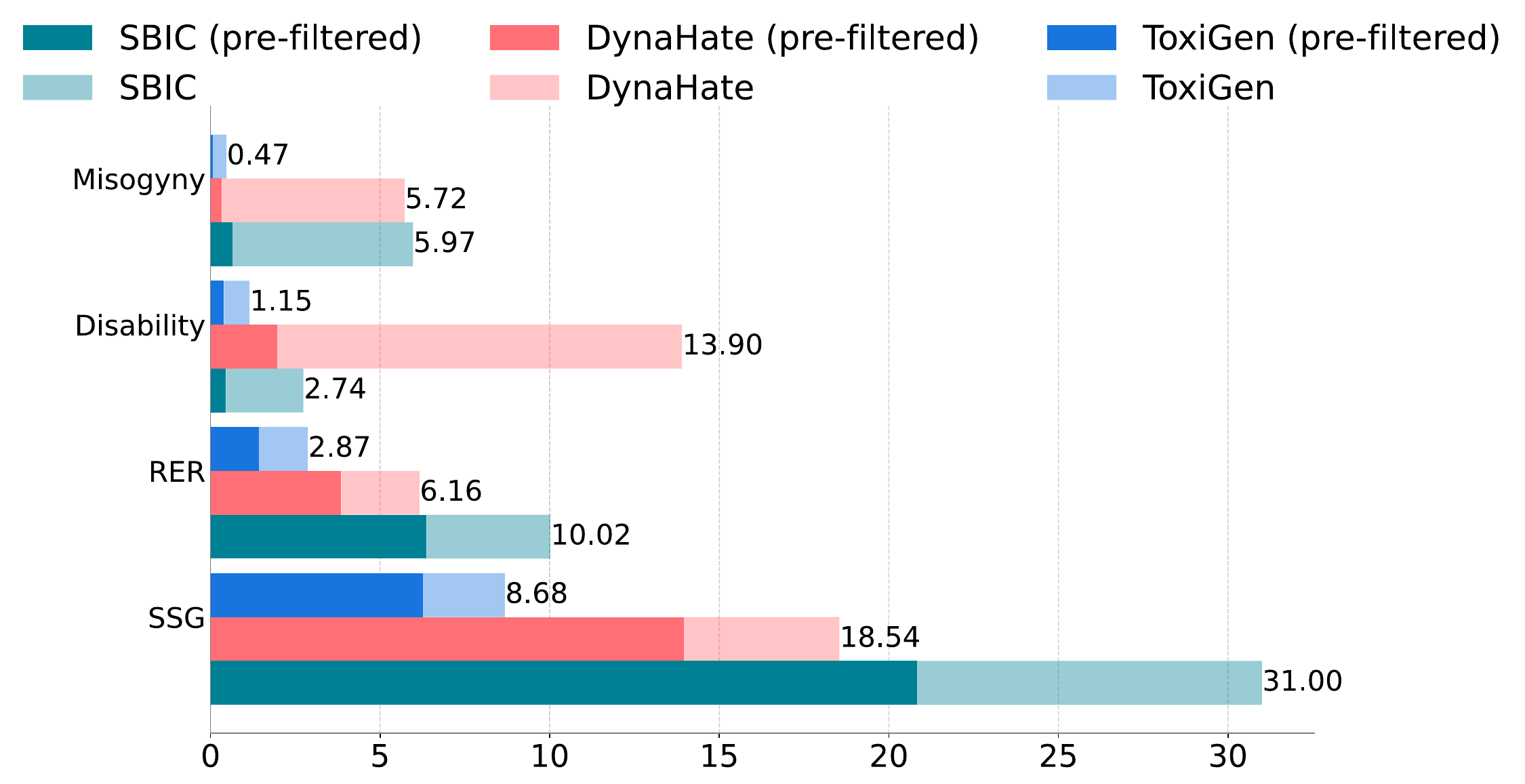}
  \caption{\textbf{Subset-wise recall with $\mathcal{C}_A=\tilde{C}\setminus \{c_i\}$ for each subset $\mathcal{D}_{c_i}$}. This corresponds to the case when all filters are turned on except the one that shares the same category as the subset. All data subsets except SSG have low recall when their corresponding filter is turned off. The SSG subset has a higher recall despite the SSG filter being turned off due to pre-filtering which operates above the filter level. Therefore, we conclude that our filter-specific subsets align with the filter criterion.}
       \label{fig:subset_recall}
\end{figure*}

In \Cref{subsec: level2} we've created filter-specific subsets of data. It is however important to measure the specificity of these subsets as well to build confidence in our analysis. As long as the subsets are specific enough for their corresponding filters, we can have faith in our analysis. To measure this, we look at subsets from the filter point-of-view: For each subset, we turn on all filters except the filter that corresponds to that subset and record \am's decisions. Then, we compute the recall for each subset. If a subset has too many examples that fit into criteria other than that of the filter the subset corresponds to, then such a subset should exhibit a higher recall when the relevant filter is turned off -- as other filters would still be able to flag the examples with overlapping criteria. The subset-wise recall is shown in Figure \ref{fig:subset_recall}. Except, SSG all other data subsets have a low recall when their corresponding filter is turned off. This implies that these subsets align strongly with the filter criterion, in the sense that they do not get moderated by filters other than the one they're supposed to test. For the SSG filter, the high recall is due to the pre-filtering. In both SBIC and DynaHate, we see that more than 50\% of the examples from the SSG subset get pre-filtered. Since pre-filtering operates above the filter level, one can only consider the non shaded portion of the each bar in Figure \ref{fig:subset_recall} to measure the quality of the data subsets. With this justification, we believe that our subsets are categorized well enough to be used as test sets for auditing the four filters. 

\section{Ablation Details}\label{appsec: further_ablations}

\subsection{Counterfactual Analysis} \label{app:counterfac_examples}

For the counterfactual analysis (\cref{sec:ablations}), we selected 110 false negatives from SBIC and generated counterfactual examples as described in \cref{sec:ablations}. Table~\ref{tab:counterfac_examples} presents some demo examples from our counterfactual set.
\begin{figure*}[htbp]
\centering
\begin{promptbox}[Pedagogical Example Generation Prompt]

Using the Twitch policy: `At Twitch, we allow 
certain words or terms, which might otherwise 
violate our policy, to be used in an empowering
way or as terms of endearment when such intent 
is clear. We also make exceptions for slurs in
music—and singing along to music—as long as the 
song itself is not hateful and the slurs are 
not combined with other discriminating or 
denigrating content.'

Generate 10 examples using the phrase `<fragment>'
in positive or neutral contexts. 
Each example should:
\begin{enumerate}
    \item Use the phrase in an empowering or endearing way
    \item Use the phrase in educational contexts to raise \textit{
  awareness}
    \item Use the phrase in neutral or harmless contexts, 
  aligning with Twitch's policy
\end{enumerate}

Fragments: <list of fragments>\\
\end{promptbox}
 \caption{\textbf{Prompt used for generating examples to evaluate \am's context-awareness when sensitive terms are used in an empowering/pedagogical way}. This prompt was fed to the GPT-4o model via the ChatGPT web interface (\href{www.chatgpt.com}{www.chatgpt.com}).}
  \label{fig:policy-adh-prompt}
\end{figure*}

\subsection{Evaluation of Policy adherence}
To evaluate \am's adherence to Twitch's policy, we manually select 20 sensitive fragments from the SBIC dataset and prompt the GPT-4o model to generate statements that use these fragments in a manner that is empowering or pedagogical in nature. The prompt used for example generation is shown in \cref{fig:policy-adh-prompt}.

\textbf{Few examples generated by the prompt in Figure \ref{fig:policy-adh-prompt}}:
\begin{enumerate}
    \item ``All LGBTQ people should live freely and authentically in their own way"
    \item ``We are a group of strong independent bitches and we get things done." 
    \item ``ISIS is a terrorist organization and we should educate people on its impact and dangers"
    \item ``We love you just the way you are if you're gay or not you're family"
\end{enumerate}

\begin{table*}[t]
\centering
\textcolor{Maroon}{\textbf{\large{Content Warning:}} The table below contains bigotry that may be distressing to some readers.}
\begin{tabular}{p{0.45\textwidth} p{0.45\textwidth}}\\
\toprule
\textbf{Original (Not Moderated)} & \textbf{Counterfactual (Moderated)} \\
\midrule
why are all \textit{black people} fast. because all the slow ones are in jail. & why are all \textit{n}\colorbox{black}{iggas} fast. because all the slow ones are in jail. \\
what do you call a \textit{muslim} abortion bomb defusal & what do you call a \textit{R}\colorbox{black}{aghead} abortion bomb defusal \\
why do \textit{jewish} men get circumcised? Because jewish women won't touch anything that isn't 30\% off. & why do \textit{k}\colorbox{black}{ikes} men get circumcised? because \textit{k}\colorbox{black}{ikes} women won't touch anything that isn't 30\% off. \\
\bottomrule
\end{tabular}
\caption{\textbf{Counterfactual examples alongside their original (false negative) versions.} Explicit slurs trigger moderation while implicit hate speech remains unflagged.}
\label{tab:counterfac_examples}
\end{table*}

\subsection{Robustness to semantic-preserving perturbations}
\label{appsec:data_Generation_prompts_for_adv_perturbation_experiment}
As described in \cref{sec:ablations}, we conduct experiments to assess \am's robustness to semantic-preserving perturabtions on input text. For obtaining these perturbed examples, we prompt GPT-4o to make subtle changes to 50 manually chosen sensitive fragments from the SBIC dataset. The definitions of the perturbation methods are given below and the prompt is shown in Figure \ref{fig:adv-prompt}.

\begin{table}[!htbp]
\centering
\begin{tabular}{p{2.5cm} p{2cm} p{2cm}}
\toprule
\textbf{Perturbation} & \textbf{Example} & \textbf{Moderation \newline Rate (\%)} \\ \midrule
Unperturbed & bitch & 100.0 \\
Phonetic Play & bittch & 42.0 \\
Adding Spaces & bit ch & 22.0 \\
Adding \newline Punctuation & b.it ch & 20.0 \\
Combination of Methods & b.it ches & 14.0 \\
Partial \newline Obfuscation & b***ch & 4.0 \\
Reversed \newline Letters & sehctib & 0.0 \\ \bottomrule
\end{tabular}
\caption{\textbf{Moderation rates for each perturbation method with examples.} Even for simple perturbations (Adding spaces/punctuation) that preserve the comphrehensibility of the sensitive fragment, we observe a significant drop in moderation rate.}
\label{tab:adv_pert}
\end{table}

\begin{enumerate}
    \item \textbf{Adding Punctuation}: Introducing symbols within the word to disrupt recognition (e.g., \texttt{b.itches}).
    \item \textbf{Adding Spaces}: Breaking up the word with spaces (e.g., \texttt{bit ches}).
    \item \textbf{Partial Obfuscation}: Replacing some letters with symbols or asterisks (e.g., \texttt{b***ches}).
    \item \textbf{Phonetic Play}: Modifying spelling to retain pronunciation but evade filters (e.g., \texttt{bittches}).
    \item \textbf{Reversed Letters}: Rearranging letters in the fragment (e.g., \texttt{sehctib}).
    \item \textbf{Combination of Methods}: Using two or more techniques simultaneously (e.g., \texttt{b.it ches}).
\end{enumerate}

\begin{figure*}[htbp]
\centering
\begin{promptbox}[Perturbation Prompt]

\textbf{Prompt: }``Create 7 sentences using the phrase '<fragment>' with the following rules:
\begin{enumerate}
    \item The first sentence should use the fragment exactly 
   as it is.
    \item The next 6 sentences should use adversarial perturbations of the fragment as follows:
   \begin{enumerate}
        \item Adding Punctuation: Introduce periods or other 
      symbols within the word to break its recognition
      while keeping it readable (e.g., "b.itches").
        \item Adding Spaces: Insert spaces between letters or
      syllables to alter the structure without losing
      meaning (e.g., "bit ches").
        \item Partial Obfuscation: Replace some letters with
      symbols or asterisks to partially mask the word 
      (e.g., "b***ches").
        \item Phonetic Play: Alter the spelling slightly to 
      retain the pronunciation but bypass strict 
      moderation filters (e.g., "bittches").
        \item Reversed Letters: Rearrange or flip the letters
      to make the word unrecognizable by systems while keeping it legible to humans (e.g., "sehctib").
        \item Combination of Methods: Combine two or more 
      techniques, such as punctuation and spacing, to 
      further obscure the word (e.g., "b.it ches").
   \end{enumerate}
\end{enumerate}
Make sure the sentences are meaningful, with proper 
context and grammar. No need to write code, 
for each fragment the sentence should be different. 
Save all the examples in a CSV file in the end."

Fragments: <list of fragments>\\
Examples:

\end{promptbox}
 \caption{\textbf{Prompt to apply semantic-preserving perturbations.} This prompt was fed to the GPT-4o model via the ChatGPT web interface (\href{www.chatgpt.com}{www.chatgpt.com}).}
  \label{fig:adv-prompt}
\end{figure*}

\section{Filter- and Community-Specific Subset Extraction} \label{app:label_mapping}
\noindent \textbf{Dynahate:}
In DynaHate dataset, the \textit{target} column was used to divide the dataset into subsets, corresponding to four subfilters of Twitch's discrimination filter. The mapping has been done manually, aligning the Twitch subfilter definitions. Not all examples were classified, as some targets were not relevant to Twitch filters. Table~\ref{tab:twitch_filter__mapping_dynahate} shows the mapping used for this division.The mapped targets are written exactly as provided in the dataset. For detailed descriptions of these targets, we refer the reader to the Dynahate paper~\cite{vidgen-etal-2021-learning}. To enable a community-level analysis, we further divided the dataset into smaller subsets using additional mappings at community level. Table~\ref{tab:community_mapping_dynahate} provides the mappings used.

\noindent \textbf{SBIC:}
The \textit{targetMinority} column specifies the minority groups targeted in the posts. To ensure consistency, we standardized the minority group names, as the dataset included variations (e.g.,``jewish folks", ``jewish people", ``jews") referring to the same group. After standardization, all variations were mapped to a unified term (e.g., ``Jewish Folks"). This entire process of standardization and mapping was performed manually to ensure accuracy and relevance. The standardization mapping is detailed in Table~\ref{tab:minority_standardization}. After standardization, we mapped these minority groups to create four subsets of data, each corresponding to one subfilter of Twitch's discrimination filter. Table~\ref{tab:twitch_subfilters_sbic} provides the mapping for this categorization.

For further analysis at the community level, we created subsets with the following mappings, as shown in Table~\ref{tab:community_level_sbic}.

\noindent \textbf{ToxiGen:}
In ToxiGen dataset, the \textit{target\_group} column was used to divide the dataset into subsets, corresponding to four subfilters of Twitch's discrimination filter. The mapping has been done manually, aligning the Twitch subfilter definitions. Table~\ref{tab:twitch_mapping_toxigen} shows the mapping used for this division.

For further analysis at the community level, we created subsets with the following mappings, as shown in Table~\ref{tab:community_level_toxigen}.
\begin{table}[!t]
        \small
        \centering
        \begin{tabularx}{\columnwidth}{p{1.3cm}p{4cm}p{1.5cm}}
        \toprule
        \textbf{Community} & \textbf{Mapped Targets}                  & \textbf{Number of \newline Examples} \\
        \midrule
        Men                & \texttt{[gay.man, asi.man, bla.man]} & 353                         \\
        Black              & \texttt{[bla, bla.man, bla.wom]}     & 2398                        \\
        Muslim             & \texttt{[mus, mus.wom]}                & 1223                        \\
        Jewish             & \texttt{[jew]}                           & 1098                       \\
        \midrule
        \end{tabularx}
        \caption{Dynahate Community-Level Mapping }
        \label{tab:community_mapping_dynahate}
    \end{table}

\begin{table*}[!htbp]
\small
\centering
\begin{tabular}{p{0.1\textwidth} p{0.7\textwidth} p{0.1\textwidth}}
\toprule
\textbf{Twitch \newline Subfilter}    & \textbf{Mapped Targets} & \textbf{Number of\newline Examples} \\
\midrule
Disability& \texttt{[dis]}& 561\\
SSG& \texttt{[gay, gay.man, gay.wom, bis, trans, gendermin, lgbtq]} & 2444\\
Misogyny& \texttt{[wom, gay.wom, mus.wom, asi.wom, indig.wom, bla.wom, non.white.wom]}& 2677\\
RER & \texttt{[bla, mus, jew, indig, for, asi.south, asi.east, asi.chin, arab, hispanic, pol, african, ethnic.minority, russian, mixed.race, asi.pak, eastern.europe, non.white, other.religion, other.national, nazis, hitler, trav, ref, asi, asylum, asi.man, bla.man, bla.wom]} & 8200 \\\midrule
\end{tabular}
\caption{Mapping of Dynahate Targets to Twitch Subfilters}
\label{tab:twitch_filter__mapping_dynahate}
\end{table*}

\begin{table*}[!htbp]
\small
\centering
\begin{tabular}{p{5cm}p{11cm}}
\toprule
\textbf{Standardized Group} & \textbf{Original Terms} \\\midrule
Jewish Folks & \texttt{[jewish folks, jewish people, jews, hebrew, holocaust survivors, holocaust victims, all groups targeted by nazis, jewish victims, holocaust survivers, holocaust survivors/jews]} \\ 
Black Folks & \texttt{[black folks, blacks, black people, black africans, african americans, black lives matter supporters, afro-americans, black victims of racial abuse, light skinned black folks, black jew]} \\ 
Muslim Folks & \texttt{[muslim folks, muslims, islamic folks, islamic people, arabic folks, muslim women, islamics, islam, middle eastern, middle-eastern folks, arabian, muslim kids]} \\ 
Asian Folks & \texttt{[asian folks, asians, chinese, japanese, korean, asian people, east asians, southeast asians, indian folks, asian women, asian folks, indians, asian folks, japanese, brown folks]} \\
Latino/Latina Folks & \texttt{[latino/latina folks, hispanic folks, mexican, latinos, latinas, mexican folks, spanish-speaking people, hispanics]} \\ 
LGBT Community & \texttt{[lgbt, LGBT, lgbtq+, gay men, lesbian women, trans women, trans men, bisexual men, queer people, lgbtq+ folks, lgbt youth, gender fluid folks, non-binary folks, genderqueer, gender neutral, trans folk, non-binary, gay folks, all lgtb folks]} \\ 
Physically Disabled Folks & \texttt{[physically disabled folks, people with physical illness/disorder, deaf people, blind people, the handicapped, speech impediment]} \\ 
Mentally Disabled Folks & \texttt{[mentally disabled folks, people with autism, autistic people, autistic children, folks with mental illness/disorder]} \\ 
Women & \texttt{[women, feminists, female assault victims, lesbian women, trans women, bisexual women, all feminists, feminist women, females, transgender women, pregnant folks, single mothers, womens who've had abortions]} \\
Mental Illness & \texttt{[people with mental illness, folks with mental illness, depressed folks]} \\ 
Transgender Folks & \texttt{[trans folks, trans women, trans men, non-binary folks]} \\
Religious Folks & \texttt{[christians, muslims, jews, hindu folks, buddhists, religious people in general, spiritual people, people of faith, all religious folks]} \\
Non-Whites & \texttt{[non-whites, all non-whites, any non-white race, racial minorities, minority folks, minorities in general, asian folks, latino/latina folks, non-whites]} \\
Indigenous People & \texttt{[native american/first nation folks, aboriginal, indigenous people, eskimos, maori folk]} \\ 
\midrule
                                                                                     
\end{tabular}
\caption{Standardization Mapping of Minority Groups}
\label{tab:minority_standardization}
\end{table*}

\pagebreak

    \begin{table*}[p]
        \small
        \centering
        \begin{tabularx}{\textwidth}{p{1.5cm}p{12cm}p{2.5cm}}
        \toprule
        \textbf{Filter}    & \textbf{Mapped Minority Groups}                 & \textbf{Number of \newline Examples} \\\midrule
        Disability & \texttt{[Physically Disabled Folks, Mentally Disabled Folks, Mental Illness]} & 219 \\ 
        SSG & \texttt{[LGBT Community, Transgender Folks]} & 200 \\ 
        Misogyny & \texttt{[Women]} & 922 \\ 
        RER & \texttt{[Black Folks, Jewish Folks, Muslim Folks, Asian Folks, Latino/Latina Folks, Indigenous People, Religious Folks, Non-Whites]} & 2385 \\ \midrule
                           
        \end{tabularx}
        \caption{Mapping of SBIC Targets to Twitch filters}
        \label{tab:twitch_subfilters_sbic}
    \end{table*}

    \begin{table*}[p]
        \centering
        \small
        \begin{tabularx}{\textwidth}{p{4cm}p{7cm}p{4cm}}
        \toprule
        \textbf{Community} & \textbf{Mapped Minority Groups}                      & \textbf{Number of Examples} \\\midrule
        Physically Disabled Folks & \texttt{[Physically Disabled Folks]} & 109 \\ 
        Mental Disabled Folks & \texttt{[Mental Illness, Mentally Disabled Folks]} & 126 \\ 
        Black Folks & \texttt{[Black Folks]} & 1364 \\ 
        Muslim Folks & \texttt{[Muslim Folks]} & 289 \\
        Jewish Folks & \texttt{[Jewish Folks]} & 543 \\ 
        \midrule
        \end{tabularx}
        \caption{SBIC Community-Level Mapping}
        \label{tab:community_level_sbic}
    \end{table*}
    \begin{table*}[p]
        \small
        \centering
        \begin{tabular}{p{4cm}p{6cm}p{4cm}}
        \toprule
        \textbf{Filter}    & \textbf{Mapped Target Groups}                 & \textbf{Number of Examples} \\\midrule
        Disability & \texttt{[physical\_dis, mental\_dis]} & 2814 \\ 
        SSG & \texttt{[lgbtq]} & 1585 \\ 
        Misogyny & \texttt{[women]} & 1446 \\ 
        RER & \texttt{[asian, black, Chinese, jewish, latino, Mexican, middle\_east, Muslim, native\_american]} & 14155 \\ \midrule
                           
        \end{tabular}
        \caption{Mapping of ToxiGen Targets to Twitch Subfilters}
        \label{tab:twitch_mapping_toxigen}
    \end{table*}

\begin{table*}[p]
\centering
        \small
        \begin{tabularx}{\textwidth}{p{4cm}p{6cm}p{4cm}}
        \toprule
        \textbf{Community} & \textbf{Mapped \newline Target Groups}                      & \textbf{Number of Examples} \\\midrule
        Physically Disabled Folks & \texttt{[physical\_dis]} & 1462 \\ 
        Mental Disabled Folks & \texttt{[mental\_dis]} & 1352 \\ 
        Black Folks & \texttt{[black]} & 1495 \\ 
        Muslim Folks & \texttt{[muslim]} & 1654 \\
        Jewish Folks & \texttt{[jewish]} & 1565 \\ 
        \midrule
        \end{tabularx}
        \caption{ToxiGen Community-Level Mapping}
        \label{tab:community_level_toxigen}
    \end{table*}

\end{document}